\documentclass{article}

\usepackage{microtype}
\usepackage{graphicx}
\usepackage{subfigure}
\usepackage{booktabs} 

\usepackage{hyperref}

\usepackage[accepted]{icml2021}

\usepackage{xcolor}
\usepackage[utf8]{inputenc}
\usepackage{makecell}
\usepackage{amsmath}
\usepackage{mathrsfs}
\usepackage{comment}
\usepackage{bm}
\usepackage{bbm}
\usepackage{amssymb}
\usepackage{amsfonts} 
\usepackage{amsmath} 
\usepackage{bbm}
\usepackage{array,multirow,graphicx}
\usepackage{etoolbox}
\usepackage{footmisc}
\usepackage{xspace}
\usepackage{booktabs}
\usepackage{soul}
\usepackage{paralist}
\usepackage{rotating}
\usepackage{adjustbox}
\usepackage[switch]{lineno}  %
\usepackage{algorithm}
\usepackage{algorithmic} 
\usepackage{tikz}
\usetikzlibrary{fit,shapes.misc}

\newcommand{\fgr}[3][\relax]{%
	\begin{figure}[htp]%
		\centering
		\includegraphics[#2]{#3}%
		\vspace{-0.1in}
		\ifx\relax#1\else\caption{{#1}}\fi
	\end{figure}%
}

\newcommand{\method}{{\sc MetaOD}\xspace}

\newcommand{\methodc}{{\sc MetaOD}{\bf\_C}\xspace}

\newcommand{\methodf}{{\sc MetaOD}{\bf\_F}\xspace}

\newcommand{\cbit}{\begin{compactitem}}
	\newcommand{\ceit}{\end{compactitem}}
\newcommand{\cben}{\begin{compactenum}}
	\newcommand{\ceen}{\end{compactenum}}

\newcommand{\bal}{\begin{align}}
\newcommand{\ean}{\end{align}}
\newcommand{\bit}{\begin{itemize}}
\newcommand{\eit}{\end{itemize}}
\newcommand{\ben}{\begin{enumerate}}
\newcommand{\een}{\end{enumerate}}
\newcommand{\beq}{\begin{equation}}
\newcommand{\eeq}{\end{equation}}

\newcommand{\R}{\mathbb{R}}

\DeclareMathAlphabet{\mathbcal}{OMS}{cmsy}{b}{n}

\newcommand{\mDt}{\mathbcal{D}_{\text{train}}}

\newcommand{\mM}{\mathbcal{M}}

\newcommand{\bXs}{\mathbf{X}_{\text{test}}}
\newcommand{\bX}{\mathbf{X}}
\newcommand{\by}{\mathbf{y}}

\newcommand{\bP}{\mathbf{P}}
\newcommand{\hP}{\widehat{\mathbf{P}}}
\newcommand{\bD}{\mathbf{D}}
\newcommand{\bM}{\mathbf{M}}

\newcommand{\bU}{\mathbf{U}}
\newcommand{\bV}{\mathbf{V}}
\newcommand{\bC}{\mathbf{C}}

\makeatletter
\renewcommand*\env@matrix[1][*\c@MaxMatrixCols c]{%
	\hskip -\arraycolsep
	\let\@ifnextchar\new@ifnextchar
	\array{#1}}
\makeatother

\newcommand{\hide}[1]{}

\newcommand{\algrule}[1][.5pt]{\par\vskip.5\baselineskip\hrule height #1\par\vskip.5\baselineskip}
\makeatother

\newcommand\marktopleft[1]{%
    \tikz[overlay,remember picture] 
        \node (marker-#1-a) at (0,1.1ex) {};%
}
\newcommand\markbottomright[1]{%
    \tikz[overlay,remember picture] 
        \node (marker-#1-b) at (0,0) {};%
    \tikz[overlay,remember picture,thick, inner sep=2pt]
        \node[draw,rectangle,fit=(marker-#1-a.center) (marker-#1-b.center)] {};%
}

\icmltitlerunning{Automating Outlier Detection via Meta-Learning}

\begin{document}

\twocolumn[
\icmltitle{Automating Outlier Detection via Meta-Learning}




\begin{icmlauthorlist}
\icmlauthor{Yue Zhao}{cmu}
\icmlauthor{Ryan A. Rossi}{adobe}
\icmlauthor{Leman Akoglu}{cmu}
\end{icmlauthorlist}

\icmlaffiliation{cmu}{Carnegie Mellon University, PA, USA}
\icmlaffiliation{adobe}{Adobe Research, CA, USA}

\icmlcorrespondingauthor{Yue Zhao}{zhaoy@cmu.edu}

\icmlkeywords{outlier detection, meta-learning}

\vskip 0.3in
]



\printAffiliationsAndNotice{}  

\begin{abstract}
Given an unsupervised outlier detection (OD) task on a new dataset, how can we automatically select a good outlier detection method and its hyperparameter(s) (collectively called a {model})?
Thus far, model selection for OD has been a ``black art''; as any model evaluation is infeasible due to the lack of ($i$) hold-out data with labels, and ($ii$) a universal objective function.  
In this work, we develop 
the {\it first principled data-driven approach to model selection for OD}, called \method,
based on meta-learning.
\method capitalizes on the past performances of a large body of detection models on existing outlier detection benchmark datasets, and carries over this prior experience to automatically select an effective model to be employed on a new dataset \textit{without using any labels}. 
To capture task similarity, we introduce specialized meta-features that quantify outlying characteristics of a dataset.
Through comprehensive experiments, we show the effectiveness of \method in selecting a detection model that significantly outperforms the most popular outlier detectors (e.g., LOF and iForest) as well as various state-of-the-art unsupervised meta-learners while being extremely fast.
To foster reproducibility and further research on this new problem, 
we open-source our entire meta-learning system, benchmark environment, and testbed datasets.

\end{abstract}

\section{Introduction}
\label{sec:intro}
The lack of a universal learning model that performs well on \textit{all} problem instances is well recognized \cite{journals/tec/DolpertM97}. Therefore, effort has been directed toward building a toolbox of various models and algorithms, which has given rise to the problem of algorithm selection and hyperparameter tuning (i.e., model selection). The same problem applies to outlier detection (OD); a long list of detectors has been developed in the last decades \cite{books/sp/Aggarwal2013}, with no universal ``winner''.
In supervised learning, model selection 
can be done via performance evaluation of each trained model on a hold-out set. When the model space becomes huge (e.g., numerous hyperparameters), basic trial-and-error techniques such as grid search become intractable. To this end, various hyperparameter optimization techniques have been developed \cite{feurer2019hyperparameter,yang2020hyperparameter} 
to strategically and iteratively search the model space based on model evaluations at carefully-chosen (hyperparameter) configurations. 
{Meta-learning} has been one of the key contributors to this effort,
thanks to the advances in gathering extensive sets of data to benchmark machine learning models. In principle, 
meta-learning is a suite of techniques that carries over past experience on a set of prior tasks to do efficient learning (e.g., fewer trial-and-errors, learning with less data, etc.) on a new task, which has been effective in automating machine learning \cite{journals/corr/abs-1810-03548}.
Different from supervised settings, unsupervised OD does not have access to hold-out data with labels, nor is there a universal objective function that could guide model selection (unlike e.g., clustering where a loss function enables model comparison). Model selection for OD is challenging exactly because model evaluation/comparison is not feasible---which renders any iterative search strategies inapplicable.
  Consequently, there has been no principled work on model selection for \textit{unsupervised} OD---the choice of a model 
  for a new task (dataset) has rather been ``a black art''.

In this work, we  tackle the {model selection} problem for outlier detection \textit{systematically}. To that end, we introduce (to the best of our knowledge) \textit{the first meta-learning based approach to OD} that selects an effective model (detector and its associated hyperparameter(s)) to be employed on a new detection task. Our proposed \method stands on the prior performances of a large collection of existing detection models on an
extensive corpora of existing outlier detection benchmark datasets. Specifically, a candidate model's performance on the new task (with no labels) is estimated based on its prior performance on {\em similar} historical tasks.

In that respect we establish a connection between the OD model selection problem and the cold-start problem in collaborative filtering (CF), where the new task is akin to a new user (with no available evaluations, hence cold-start) and the model space is analogous to the item set.
Differently, OD necessitates the identification of a single best model (i.e., top rank or single-shot selection), whereas CF typically operates in a top-$k$ setting. In CF, future recommendations can be improved based on user feedback which is not applicable to OD.
Moreover, 
\method requires
the effective learning of task similarities based on characteristic dataset features (namely, meta-features) that capture the outlying properties within a dataset, whereas user features  (location, age, etc.) in CF may be readily available.

In summary, the key contributions of this work include:

\cbit
    \item \textbf{The First OD Meta-learner}: We propose \method, (to our knowledge) the first systematic effort on model selection for OD, which stands on \textit{unsupervised} meta-learning in principle, and historical collections of trained models and benchmark datasets in practice.
    \item\textbf{Problem Formulation:} We establish a link with CF under cold-start, where the new task ``better likes'' a model that performs better on similar past tasks.
    
    \item\textbf{Specialized Meta-features for OD:} We design novel meta-features to  capture the outlying characteristics in a dataset toward effectively quantifying task similarity. 
    \item \textbf{Effectiveness and Efficiency}:
     Given a new dataset, \method selects a detector and its associated hyperparameter(s) to be employed in a \textit{single-shot} (i.e., without requiring any model evaluations), while incurring negligible run-time overhead. Through  extensive experiments on two  benchmark testbeds that we have constructed, we show that \textit{selecting} a model by \method significantly outperforms state-of-the-art unsupervised meta-learners and various popular models like iForest.
    \item \textbf{Open-source Platform}: We open-source 
    \method
    and our meta-learning database\footnote{\label{code_url}Anonymous URL: \url{http://bit.ly/MetaOD}}
    for the community to deploy it 
    for OD model selection and to extend the database with new datasets and models. 
    We expect the growth of the database would make OD meta-learning more powerful  and also help foster 
    further research on this new problem. 
\ceit

\section{Related Work}
\label{sec:related_works}

\subsection{Model Selection for Outlier Detection (OD)}
Most  outlier  detection  work  have  focused  on  developing  better  methods  for  different  types  of data and problem settings \cite{books/sp/Aggarwal2013}.
What has not been tackled thus far is the general OD model selection problem---\textit{which} method to use on a new task. The challenges are two-fold: 
lack of access to ground truth labels, 
and an established OD loss function; rendering model evaluation and comparison inapplicable.

We note that {\em some} OD methods have a loss function;  auto-encoders  \cite{chen2017outlier,zhou2017anomaly} use reconstruction error and one-class classification (OCSVM \cite{scholkopf2000support}, SVDD \cite{tax2004support}) aims to maximize 
margin to origin or minimize the radius of data-enclosing hyperball.
There exist some work on model selection 
for one-class models  \cite{burnaev2015model,xiao2014two,evangelista2007some,Deng07}, however, those \textit{apply only to this specific model class} and not in the general case.
Our proposed \method is not limited to one model class, but 
can select among any (heterogeneous) set of methods.

\vspace{-0.1in}
\subsection{Model Selection, AutoML, Meta-Learning}
Model selection refers to the process of algorithm selection and/or hyperparameter optimization (HO). 
With the advance of complex (e.g., deep) models, 
HO in high dimensions \cite{yu2020hyperparameter} has become impractical to be human-powered. As such, automating ML pipelines has seen a surge of attention \cite{he2019automl}.  
Meta-learning \cite{journals/corr/abs-1810-03548,elshawi2019automated,yao2018taking} has been a key contributor to the effort, 
which aims to design models for new tasks based on prior experience. 


\vspace{-0.075in}
\subsubsection{Supervised Model Selection}
Most existing work focus on the supervised setting, which leverage hold-out data with labels.
Randomized \cite{bergstra2012random}, bandit-based \cite{li2017hyperband}, and 
Bayesian optimization (BO) techniques \cite{shahriari2015taking} are various SOTA approaches to HO.
Specifically sequential model-based BO \cite{journals/jgo/JonesSW98,hutter2011sequential} evaluates hold-out performance 
at various initial HCs, a (smooth) surrogate meta-function is fit to the resulting (HC, performance) pairs, which is then used to strategically query other HCs, e.g., via hyper-gradient based search \cite{franceschi2017forward}. The query function needs to carefully tradeoff exploration of new promising HCs and exploitation via local search around well-performing HCs.
Meta-learning is used to find promising 
initialization for (i.e., warm-starting) BO \cite{conf/dsaa/WistubaSS15,conf/aaai/FeurerSH15,journals/corr/abs-1802-02219,journals/ml/WistubaSS18}.

Note that \textit{all of these approaches rely on multiple model evaluations} (i.e., performance queries) at different HCs, and hence cannot be applied to OD model selection problem.

\vspace{-0.075in}
\subsubsection{Unsupervised Model Selection}
Unsupervised ML tasks (e.g., clustering)
poses additional challenges \cite{vaithyanathan2000generalized,fan2019unified}.
Nonetheless, those exhibit \textit{established objective criteria} that enable model comparison, \textit{unlike} OD. For example, BO methods still apply where the surrogate can be trained on (HC, objective value) pairs, for which meta-learning can provide favorable priors. 

Task-independent meta-learning \cite{abdulrahman2018speeding}, which simply identifies the globally best model on historical tasks, applies to the unsupervised setting as well as OD. This can be refined by identifying the best model on {\em similar} tasks, where task similarity is measured in the meta-feature space via clustering \cite{conf/ecai/KadiogluMST10} or nearest neighbors \cite{nikolic2013simple}.

This type of similarity-based recommendations points to a connection between algorithm selection and collaborative filtering (CF), first recognized by \cite{stern2010collaborative}. 
The most related 
to OD model selection is CF under {\em cold start} (CFCS), where
evaluations are not-available (in our case, infeasible) for a new user or item (in our case, task).
There have been a number of work using meta-learning for solving the cold-start recommendation problem~\cite{bharadhwaj2019meta,lee2019melu,conf/nips/VartakTMBL17}, and vice versa, using CFCS solutions for 
ML algorithm selection \cite{journals/ai/MisirS17}. The latter 
is most related to our work, to which we compare in experiments.

\section{Meta-Learning for OD Model Selection}
\label{sec:method}
\subsection{Problem Statement}

We tackle the {model selection problem for unsupervised outlier detection systematically via meta-learning}.
Being a meta-learning based approach, the proposed \textit{meta-learner} \method relies on 
\cbit
\item
a collection of historical outlier detection datasets 
$\mDt = \{\bD_1,\ldots,\bD_n\}$, namely, a meta-train database with ground truth labels, i.e., $\{\bD_i = (\bX_i,\by_i)\}_{i=1}^n$,
\vspace{0.1in}
\item as well as  historical performances of the models that define the model space, denoted $\mM = \{M_1, \ldots, M_m\}$, on the meta-train datasets.
\ceit
We refer to $\bP\in \R^{n\times m}$ as the performance matrix, where $\bP_{ij}$ corresponds to the $j$th model $M_j$'s performance\footnote{Area under the precision-recall curve (a.k.a. Average Precision or AP); can be substituted with any other  measure of interest.}  on the $i$th meta-train dataset $\bD_i$.\footnote{Model evaluation \textit{is} possible on meta-train datasets as they contain ground-truth labels. This is not the case at test time.}

The OD model selection problem is then stated as:
\textit{Given} a new input dataset $\bXs$ (i.e., detection task) with \textit{no} labels, \textit{Select} a model $M \in \mM$ to employ on the new (test) task.

{\bf Remark:} Model selection for OD involves selecting both ($i$) a detector/algorithm (discrete) and ($ii$) its associated hyperparameter(s) (continuous). Due to the latter, the model space in fact consists of infinitely many models. 
Under certain assumptions, such as performance changing smoothly in the hyperparameter space, a hyperparameter configuration can be selected \textit{iteratively} based on evaluations on carefully-chosen prior configurations.
Importantly, we remark that OD is not amenable for such iterative search over models---evaluations are \textit{not} possible due to the lack of both labels and an objective criterion. The selection of a model, therefore, is to be done in a \textit{single-shot}.
Given this limitation that stems from the problem domain, we discretize the hyperparameter space for each detector to make the overall search space tractable.
As such, each model $M \in \mM$ can be thought as a $\{$detector, configuration$\}$ pair, for a specific configuration of the hyperparameter(s). (See Appendix \ref{appendix:model_set}.)

\subsection{Proposed \method}
\label{subsection:system_overview}

Our \method consists of two-phases: \textbf{offline training} of the meta-learner and \textbf{online prediction} that yields model selection at test time. 
Arguably, the running time of the offline phase is not critical. In contrast, model selection for an input task should incur very small run-time overhead, as it precedes the actual building of the selected OD model.

\hide{
 \begin{table}[!t]
 \footnotesize
 \begin{center}
 \begin{tabular}{ l  l l}
 \hline 
 {\bf Symbols} & {\bf Shape} & {\bf Definitions} \\ 
 \hline
 \hline
 $n$ & scalar & \# of training datasets \\ 
 $d$ & scalar & \# of models to select from \\ 
 $l$ & scalar & \# meta-features (equal to 200)\\
 $P$ & $(n,d)$ & Performance matrix (model performance) \\
 $k$ & scalar & \# of latent dim. for $P$ matrix factorization \\
 $U$ & $(n,k)$& Data matrix (sub-matrix of $P$) \\ 
 $V$ & $(d,k)$& Model matrix (sub-matrix of $P$)\\
 \hline
 \end{tabular} 
 \caption{Symbols and Definitions \label{table:symbols}}
 \end{center}
 \normalsize
 \end{table}
}

\subsubsection{Offline (meta-)training:} 
\label{subsubsec:offline_training}
In principle, meta-learning carries over prior experience on a set of historical tasks to ``do better'' on a new task. Such improvement can be unlocked only if the new task \textit{resembles}  and thus can build on \textit{at least some} of the historical tasks (such as learning ice-skating given prior experience with roller-blading), rather than representing completely unrelated phenomena. 
This entails defining an effective way to capture task similarity between an input task and the historical tasks at hand.

In machine learning, similarity between meta-train and test datasets are quantified through characteristic features of a dataset, also known as {\em meta-features}. Those typically capture statistical properties of the data distributions. (See survey \cite{journals/corr/abs-1810-03548} for various types of meta-features.)

To capture \textbf{prior experience}, \method first constructs the performance matrix $\bP$ by running/building and evaluating all the $m$ models in our defined model space on all the $n$ meta-train datasets.\footnote{Note that this step takes considerable compute-time, which however amortizes to ``do better'' for future tasks. To this effect, we open-source our \textit{trained} meta-learner to be readily deployed. 
}
To capture \textbf{task similarity}, it then extracts a set of $d$ meta-features from each meta-train dataset,  denoted by $\bM=\psi(\{\bX_1, \ldots, \bX_n\})\in \mathbbm{R}^{n \times d}$ where $\psi(\cdot)$ depicts the feature extraction module. 
We defer the details on the meta-feature specifics to \S \ref{subsection:meta_feature_generation}.

At this stage, it is easy to recognize the  connection between the unsupervised OD model selection and the collaborative filtering (CF) under cold start problems.
Simply put, meta-train datasets are akin to existing users in CF that have prior evaluations on a set of models akin to the item set in CF. The test task is akin to a newcoming user with no prior evaluations (and in our case, no possible future evaluation either), which however exhibits some pre-defined features.

 Capitalizing on this connection, we take a matrix factorization based approach where $\bP$ is approximated by the dot product of what-we-call dataset matrix $\bU \in  \mathbbm{R}^{n \times k}$ and model matrix $\bV \in  \mathbbm{R}^{m \times k}$. The intent is to capture the inherent dataset-to-model affinity via the dot product similarity in the $k$-dimensional latent space, such that $\bP_{ij} \approx \bU_i{\bV_j}^T$ where matrix subscript depicts the corresponding row.

 What loss criterion is suitable for the factorization?
 In CF the typical goal is top-$k$ item recommendation. In \method, we aim to select the best model with the largest performance on a task which demands top-$1$ optimization. 
 Therefore, we discard least squares and instead optimize the rank-based (row- or dataset-wise) discounted cumulative gain (DCG), 
 
\begin{equation}
     \max_{\bU, \bV} \;\; \sum_{i=1}^{n} \text{DCG}_i(\bP_i, \bU_i \bV^T) \;. \label{equ:objective}
\end{equation}
The factorization is solved iteratively via alternating optimization, where initialization plays an important role for such non-convex problems. We find that initializing $\bU$, denoted $\bU^{(0)}$, based on meta-features facilitates stable training, potentially by hinting at inherent similarities among datasets as compared to random initialization.
Specifically, an embedding/dimensionality reduction technique $\phi$ is used to set $\bU^{(0)} := \phi(\bM)$ for $\phi: \R^d \mapsto \R^k$, $k<d$.
Details on objective criteria and optimization are deferred to \S \ref{subsection:PMF}.

By construction, matrix factorization is transductive. On the other hand, we would need $\bU_\text{test}$ in order to be able to estimate performances of the model set on a new dataset $\bXs$. To this end, one can learn an (inductive) multi-output regression model that maps the meta-features onto the latent features.  We simplify by learning a regressor $f: \R^k \mapsto \R^k$ that maps the (lower dimensional) embedding features $\phi(\bM)$ (which are also used to initialize $\bU$) onto the final optimized $\bU$. Note that this requires an inductive embedding function $\phi(\cdot)$ to be applicable to newcoming datasets.
In implementation, we use PCA for $\phi(\cdot)$ and a random forest regressor for $f(\cdot)$ although \method is flexible to accommodate any others provided they are inductive.

\subsubsection{Online prediction and model selection:} 
\label{subsubsec:prediction_model_selection}
Model selection at test time does \textit{not use any labels}.
Given a new dataset $\bXs$ for OD, \method first computes the corresponding meta-features as $\bM_{\text{test}}:=\psi(\bXs) \in \mathbbm{R}^{d}$. Those are then embedded via $\phi(\bM_{\text{test}}) \in \R^k$, which are regressed to obtain the latent features, i.e., $\bU_{\text{test}}:=f(\phi(\bM_{\text{test}})) \in \R^k$. 
Model set performances are predicted as $\bP_{\text{test}} := \bU_{\text{test}} \bV^T \in \R^m$. The model with the largest predicted performance is output as the selected model, i.e.,
\begin{equation}
\label{eq:test}
\arg\max_j  \;\; \langle f(\phi(\psi(\bXs))), {\bV_j} \rangle \;.
\end{equation}

{\bf Remark:} Note that model selection 
by \eqref{eq:test}
does not rely on ground-truth labels, hence \method provides \textit{unsupervised} outlier model selection. 
In terms of computation, test-time embedding and regression by pre-trained models $\phi$ (PCA) and $f$ (regression trees) take near-constant time given the small number of meta-features, embedding dimensions, and trees of fixed depth.  
Moreover, we rely on meta-features with computational complexity linear in the dataset size.

\subsection{Meta-Features for Outlier Detection}
\label{subsection:meta_feature_generation}
A key part of \method is the extraction of meta-features that capture the important characteristics of an arbitrary dataset. 
Existing outlier detection models have different methodological designs  (e.g., density, distance, angle, etc. based) and different assumptions around the topology of outliers (e.g., global, local, clustered). As a result, we expect different models to perform differently depending on the type of outliers a dataset exhibits.
Given a new dataset, the goal is to identify the datasets in the meta-train database that exhibit \textit{similar} characteristics and focus on models that do well on those datasets. This is akin to recommending to a new user those items liked by similar users.

To this end, we extract meta-features that can be organized into two categories:
(1) statistical features, and (2) landmarker features.
Broadly speaking, the former captures statistical  properties of the underlying data distributions; e.g., min, max, variance, skewness, covariance, etc. of the features and  feature combinations. (See Appendix \ref{appendix:od_meta_features} Table~\ref{table:meta_features} for the complete list.) These kinds of meta-features have  been commonly used in the AutoML literature.

The optimal set of meta-features has been shown to be application-dependent~\cite{journals/corr/abs-1810-03548}. 
Therefore, perhaps more important are the landmarker features, which are \textit{problem-specific}, and aim to capture the \textit{outlying} characteristics of a dataset.
The idea is to apply a few of the fast, easy-to-construct OD models on a dataset and extract features from ($i$) the structure of the estimated OD model, and ($ii$) its output outlier scores.
For the OD-specific landmarkers, we use four OD algorithms: iForest \cite{liu2008isolation}, HBOS \cite{goldstein2012histogram}, LODA \cite{pevny2016loda}, and PCA \cite{shyu2003novel} (reconstruction error as outlier score).
Consider iForest as an example. It creates a set of what-is-called extremely randomized trees that define the model structure, from which we extract structural features such as average horizontal and vertical tree imbalance.
As another example, LODA builds on random-projection histograms from which we extract features such as entropy.
On the other hand, based on the list of outlier scores from these models, we compute features such as dispersion, max consecutive gap in the sorted order, etc.
We elaborate on the details of the landmarker features in Appendix \ref{appendix:landmarker}.

\renewcommand{\algorithmicrequire}{\textbf{Input:}}
\renewcommand{\algorithmicensure}{\textbf{Output:}}
\renewcommand{\algorithmiccomment}[1]{\hfill$\blacktriangleright$ #1}
\newcommand{\cdash}{\multicolumn{1}{c}{--} }

\begin{algorithm}[t]
	\caption{\method: Offline and Online Phases}
	\label{algo:metod}
	\begin{algorithmic}[1]
		\small{
		\REQUIRE (Offline) meta-train database $\mDt$, model set $\mM$, latent dimension $k$; (Online) new OD dataset $\bXs$
		\ENSURE (Offline) Meta-learner for OD model selection; (Online) Selected model for $\bXs$
		\algrule
		
		\textcolor{blue}{$\blacktriangleright$ (Offline) OD Meta-learner Training} {{\textcolor{blue}{\ttfamily{\bf (\S \ref{subsubsec:offline_training})}}}}
		\STATE Train \& evaluate $\mM$ on $\mDt$ to get performance matrix $\bP$
		\STATE Extract meta-features {{\textcolor{blue}{\ttfamily{\bf (\S \ref{subsection:meta_feature_generation})}}}}, $\bM:=\psi(\{\bX_1, \ldots, \bX_n\})$
		\STATE Init. $\bU^{(0)}$ by embedding meta-features,  $\bU^{(0)}:=\phi(\bM; k)$
		\STATE Init. $\bV^{(0)}$ by standard normal dist.n $\bV^{(0)} \sim \mathcal{N}(0,1) $
		\WHILE[alternate. opt. by SGD, {{\textcolor{blue}{\ttfamily{\bf (\S \ref{subsection:PMF})}}}}]{not converged}
		\STATE Shuffle dataset order in $\mDt$
		\FOR{$i = 1,\ldots,n$}
		\STATE Update $\bU_i$ by Eq. (\ref{eq:gradient_U})
		\FOR{$j = 1,\ldots,m$}
		\STATE{Update $\bV_j$ by Eq. (\ref{eq:gradient_V})}
		\ENDFOR
		\ENDFOR
		\ENDWHILE
		\STATE Train $f$ regressing $\phi(\bM; k)$ onto $\bU$ (at convergence)
		\STATE \textbf{Save} extractors $\psi$, embed. $\phi$, regressor $f$, $\bV$ (at conv.)
	    \algrule
		
		\textcolor{blue}{$\blacktriangleright$ (Online) OD Model Selection} {{\textcolor{blue}{\ttfamily{\bf (\S \ref{subsubsec:prediction_model_selection})}}}}
		\STATE Extract meta-features, $\bM_{\text{test}}:=\psi(\bXs)$
		\STATE Get latent vector after embedding, $\bU_{\text{test}}:=f(\phi(\bM_{\text{test}}))$
		\STATE Predict model set performance, $\bP_{\text{test}} := \bU_{\text{test}} \bV^T $
        \STATE \textbf{Return} $\arg\max_j \bP_{\text{test}}(j)$ as the selected model for $\bXs$
	}
	\end{algorithmic}
\end{algorithm}
\setlength{\textfloatsep}{0.25in}
\pagebreak

\subsection{Meta-Learning Objective and Training}
\label{subsection:PMF}
In this section we provide details regarding our matrix factorization objective function in Eq. \eqref{equ:objective} and its optimization.

\subsubsection{Rank-based Criterion}

A typical loss criterion for matrix factorization is the mean squared error (MSE), a.k.a. the Frobenius norm of the error matrix $\bP-\bU\bV^T$. While having nice properties from an optimization perspective, MSE does not (at least directly) concern with the ranking quality. In contrast, our goal is to rank  the models  {for \textit{each} dataset} row-wise, as model selection concerns with picking the best possible model to employ.
Therefore, we use a rank-based criterion from the information retrieval literature called DCG. For a given ranking, DCG is defined as
\begin{equation}
\text{DCG} = \sum_r \frac{b^{rel_r}-1}{\log_2(r+1)} \label{equ:DCG_stronger}
\end{equation}
\noindent
where 
$rel_r$ depicts the relevance of the item 
ranked at the $r$th position and $b$ is a scalar (typically set to 2). 
In our setting, we use the performance of a model to reflect its relevance to a dataset. As such, DCG for dataset $i$ is re-written as
 \begin{align} \label{eq:DCG_h}
 \text{DCG}_i
 &= \sum_{j=1}^m \frac{b^{{\bP_{ij}}}-1}
 {\log_2(1+\sum_{k=1}^m \mathbbm{1}{[\hP_{ij} \leq \hP_{ik}}])}
 \end{align}
where $\hP_{ij} = \langle \bU_i, \bV_j \rangle$ is the predicted performance.
Intuitively, ranking high-performing models at the top leads to higher DCG, and a larger $b$ increases the emphasis on the quality at higher rank positions. 

A challenge with DCG is that it is not differentiable, unlike MSE, as it involves ranking/sorting.. 
Specifically, the sum term in the denominator of Eq. (\ref{eq:DCG_h}) uses (nonsmooth) indicator functions to obtain the position of model $j$ as ranked by the estimated performances. 
We circumvent this challenge by replacing the indicator function by the (smooth) sigmoid approximation  \cite{frery2017efficient} as follows.
{
\begin{equation}\label{eq:indicator_approx}
\text{DCG}_i  \approx \text{sDCG}_i 
= \small{\sum_{j=1}^m \frac{b^{{\bP_{ij}}}-1}{\log_2(1+\sum_{k=1}^m \sigma(\hP_{ik} - \hP_{ij}))}}
\end{equation}
}
\vspace{-0.1in}

\subsubsection{Initialization \& Alternating Optimization}
We optimize the smoothed loss 
\beq
L = - \sum_i \text{sDCG}_i(\bP_i, \bU_i\bV^T)\eeq 
by alternatingly solving 
for $\bU$ 
as we fix $\bV$ (and vice versa) by gradient descent.
We initialize $\bU$ by leveraging the meta-features, which are embedded to a space with the same size as $\bU$.
By capturing the latent similarities among the datasets, such an initialization not only accelerates convergence \cite{zheng2007initialization} but also facilitates convergence to a better local optimum.
$\bV$ is initialized via random draw from a unit Normal.

As we aim to maximize the total \textit{dataset-wise} DCG, we make a pass over meta-train datasets one by one at each epoch as shown 
in Algorithm~\ref{algo:metod}. 
For brevity, we give the gradients for $\bU_i$ and $\bV_j$ in Eq.s (\ref{eq:gradient_U}) and (\ref{eq:gradient_V}), respectively. 
\setlength{\belowdisplayskip}{0pt} \setlength{\belowdisplayshortskip}{0pt}
{\scriptsize{
\begin{align} \label{eq:gradient_U}
  \frac{\partial L} {\partial \bU_i} 
&= \ln{(2)} \sum_{j=1}^m \left[\frac{b^{\bP_{ij}}-1}{\beta^i_{j} \ln ^2{(\beta^i_{j})}} \sum_{k \neq j} \sigma(w^i_{jk})(1-\sigma(w^i_{jk}))(\bV_k-\bV_j)\right]
\end{align}
}}
\setlength{\belowdisplayskip}{0pt} \setlength{\belowdisplayshortskip}{0pt}
{\scriptsize{
\begin{align} \label{eq:gradient_V}
\frac{\partial L} {\partial \bV_j} 
&= -\ln{(2)} \sum_{j=1}^m \left[\frac{b^{\bP_{ij}}-1}{\beta^i_{j} \ln ^2{(\beta^i_{j})} } {\sum_{k \neq j} \sigma(w^i_{jk})(1-\sigma(w^i_{jk}))\bU_i}\right]
\end{align}
}}

\noindent
where $w^i_{jk}=\langle \bU_i,  (\bV_k-\bV_j) \rangle$ and $\beta^i_j = \frac{3}{2}+\sum_{k \neq j} \sigma (w^i_{jk})$.
We provide the detailed derivations in Appendix \ref{appendix:gradient}.

The steps of \method, for both meta-training (offline) and model selection (online), are given in Algorithm \ref{algo:metod}. 
Appendix \ref{sec:flow} illustrates the workflow with a diagram.

\section{Experiments}
\label{sec:experiments}
We evaluate \method by designing experiments to answer the following research questions:
\cben
    \item Does employing \method for model selection yield improved detection performance, as compared to \textit{no} model selection, as well as other selection techniques adapted from the unsupervised AutoML? (\S\ref{sec:exp-poc-testbed})
    
    \item How does the extent of similarity of the test dataset to the meta-train affect \method's performance (i.e., the relevance of the new task to prior experience)? (\S\ref{sec:exp-st-testbed})
    
    \vspace{0.01in}
    \item How much run-time overhead does \method incur preceding the training of the selected model?
    (\S\ref{sec:exp-runtime})

\ceen

\subsection{Experiment Setting}
\subsubsection{Models and Evaluation}
By pairing eight state-of-the-art (SOTA) OD algorithms and their corresponding hyperparameters, we compose a model set $\mM$ with 302 unique models. 
(See Appendix \ref{appendix:model_set} Table \ref{table:models_long} for the complete list). 
For both testbeds introduced below, we first generate the  performance matrix $\bP$, by evaluating the models from $\mM$ against the benchmark datasets in a testbed. 
As various detectors employ randomization (e.g., random-split trees, random projections, etc.), we run five independent trials and record the averaged performance. For consistency, all  models are built using the PyOD library \cite{zhao2019pyod} on an Intel i7-9700 @3.00 GHz, 64GB RAM, 8-core workstation. 
We evaluate \method and the baselines via cross-validation where each fold consists of hold-out datasets treated as test.
To compare two methods statistically, we use  
the pairwise Wilcoxon rank test on performances across datasets (significance level $p<0.05$).

\begin{figure}[!t]
\centering
    \includegraphics[width=0.7\columnwidth]{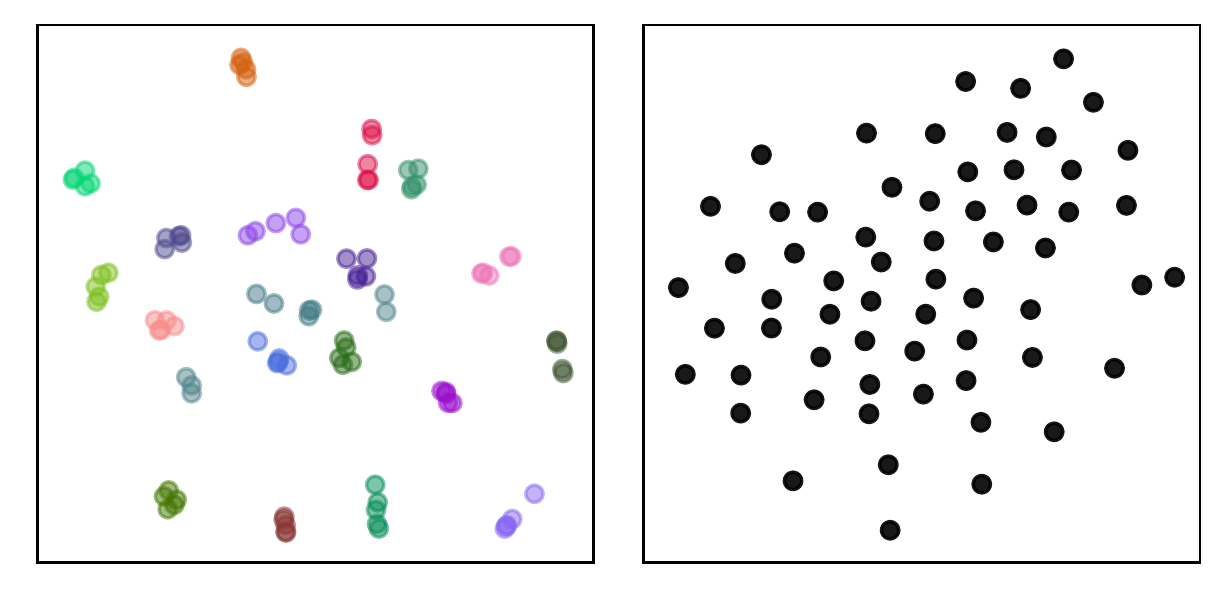}
    \vspace{-0.15in}
\caption{
t-SNE embedding of datasets in (left) POC and (right) ST. POC exhibits higher task similarity; ``siblings'' (marked by same color) form clusters. ST contains independent datasets with no apparent clusters.}
\label{fig:meta_feature}
\vspace{-0.15in}
\end{figure}

\vspace{-0.025in}
\subsubsection{Testbed Setup}
Meta-learning works if the new task can leverage prior knowledge; e.g., mastering motorcycle can benefit from bike riding experience. 
As such, \method relies on the assumption that the incoming test datasets share similarity with the meta-train datasets.
Otherwise, the learned knowledge from meta-train cannot be transferred to new datasets.
We therefore create two testbeds with varying extent of similarity among the train/test datasets to illustrate the aforementioned point:
\cben
    \item \textbf{Proof-of-Concept (POC) Testbed} has 100 datasets that form clusters of similar datasets. 
    We achieve this by creating 5 different detection tasks (``siblings'') from each of the 20 ``mother'' datasets or mothersets. 
    \item \textbf{Stress Testing (ST) Testbed} 
    consists of 62 independent datasets from public repositories, which exhibit lower similarity to one another.
\ceen

We refer to Appendix \ref{appendix:databases} for the complete list of datasets and details on testbed generation. Fig. \ref{fig:meta_feature} illustrates the differences between POC and ST, where the meta-features of their constituting datasets are t-SNE embedded to 2-D. By construction, POC consists of clusters and hence exhibits higher task/dataset similarity as compared to ST. 

\subsubsection{Baselines}
\label{subsection:baseline}
Being the first work for OD model selection, we do not have immediate baselines for comparison. Therefore we adapt leading methods from algorithm selection, and include additional baselines by creating variations of the proposed \method 
(marked with $\dagger$).
All the baselines can be organized into three categories:

\noindent\textit{\textbf{No model selection}} always employs either the same single model or the ensemble of all the models: 
\begin{compactitem}
\item \textbf{Local outlier factor (LOF)} \cite{breunig2000lof} is a popular OD method that measures a sample's deviation in the local region regarding its neighbors. 
\item \textbf{Isolation Forest (iForest)} \cite{liu2008isolation} is a SOTA tree ensemble that measures the difficulty of ``isolating" a sample via randomized splits in feature space.
\item \textbf{Mega Ensemble (ME)} averages outlier scores from the 302 models for a given dataset. ME does not perform model selection but rather uses \textit{all} the models.
\end{compactitem}

\noindent\textit{\textbf{Simple meta-learners}} pick the generally well-performing model, globally or locally:
\begin{compactitem}
\item \textbf{Global Best (GB)} is the \textit{simplest meta-learner} that selects the model with the largest avg. performance across all train datasets, \textit{without} using meta-features.
\item \textbf{ISAC} ~\cite{conf/ecai/KadiogluMST10} clusters the meta-train datasets based on meta-features. Given a new dataset, it identifies its closest cluster and selects the best model with largest avg. performance on the cluster's datasets. 
\item \textbf{ARGOSMART (AS)} ~\cite{nikolic2013simple} finds the closest meta-train dataset (1NN) to a given test dataset, based on meta-feature similarity, and selects the model with the best performance on the 1NN dataset.

\end{compactitem}

\noindent \textit{\textbf{Optimization-based meta-learners}} {\em learn} meta-feature by task similarities toward optimizing performance estimates:

\begin{compactitem}
\item \textbf{Supervised Surrogates (SS)} \cite{xu2012satzilla2012}: 
Given the meta-train datasets, 
it directly maps the meta-features onto model performances by regression.
\item \textbf{ALORS} \cite{journals/ai/MisirS17} factorizes the performance matrix to latent factors, 
and estimates performance as dot product of the latent factors. A non-linear regressor maps meta-features onto latent factors.
\item $\dagger${\methodc} is a variant: performance and meta-feature matrices are concatenated as $\bC=[\bP, \bM]
 \in \mathbbm{R}^{n\times (m+d)}$, before factorization, $\bC \approx \bU\bV^T$.
Given a new dataset, zero-concatenated meta-features are projected and reconstructed as $[\hP_{\text{new}}; \widehat{\bM}_{\text{new}}] =  [0\ldots 0;{\bM}_{\text{new}}] \bV\bV^T$.

\item $\dagger${\textbf{\methodf}} is another variant, where $\bU$ is fixed at $\phi(\bM)$ after the embedding step; only $\bV$ is optimized. 

\end{compactitem}

\vspace{0.025in}
In addition, we report \textbf{Empirical Upper Bound (EUB)} (only applicable to POC): Each POC dataset has 4 ``siblings" from the same motherset with similar outlying properties. We consider the performance of the best model on a dataset's ``siblings" as its EUB, as siblings provide significant information as to which models are suitable. Given the lower task similarity in ST testbed (that is challenging for meta-learning), we include \textbf{Random Selection (RS)} baseline to quantify how the methods compare to random, and report the smallest $q$ that \method does not significantly differ from.

\subsection{POC Testbed Results} \label{sec:exp-poc-testbed}

\subsubsection{Testbed Setting}
\label{subsec:POC}
POC testbed is built to simulate when there are similar meta-train tasks to the test task. 
We use the benchmark datasets\footnote{\url{https://ir.library.oregonstate.edu/concern/datasets/47429f155}} by \cite{Emmott2016AnomalyDM}, which created ``childsets" from 20 independent ``mothersets" (mother datasets) by sampling.
Consequently, the childsets generated from the same motherset with the same generation properties, e.g., the frequency of anomalies, are deemed to be ``siblings" with large similarity. We build the POC testbed by using 5 siblings from each motherset, resulting in 100 datasets. We split them into 5 folds for cross-validation. Each test fold contains 20 independent childsets without siblings.

\begin{figure}[!t]
\centering
 \includegraphics[scale=0.4]{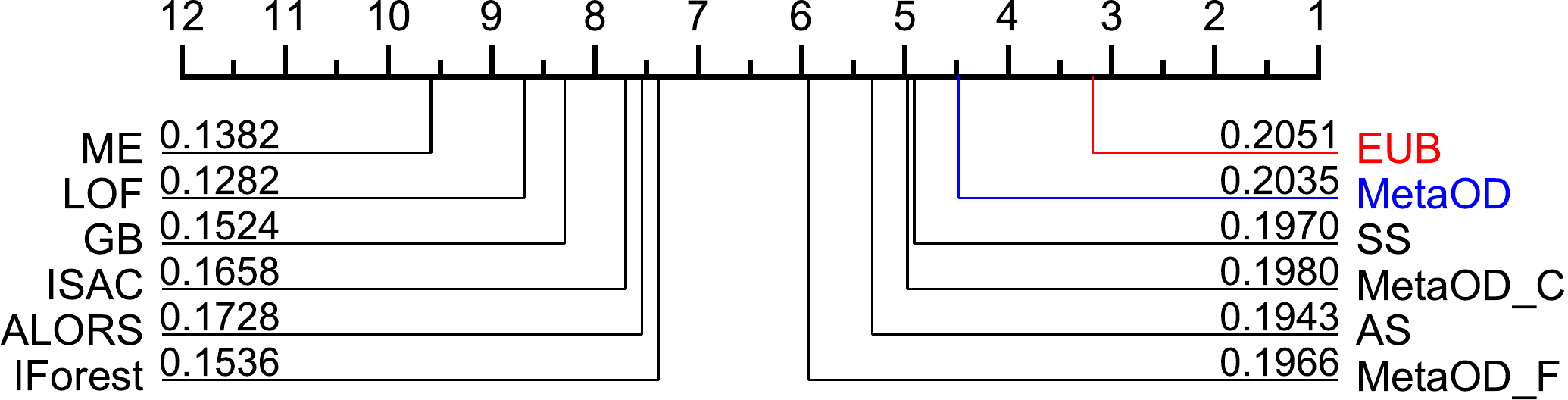}
    \vspace{-0.25in}
    \caption{
    Comparison of avg. rank (lower is better) of methods w.r.t. performance across datasets in POC. 
    Mean AP across datasets (higher is better) shown on lines. \method is the best performing meta-learner, and comparable to EUB. 
} 
\label{fig:CI_POC}
\vspace{-0.1in}
\end{figure}

\begin{table}[!htp]
\scriptsize
\caption{
Pairwise statistical test results between \method and baselines by Wilcoxon signed rank test. Statistically better method shown in \textbf{bold} (both marked \textbf{bold} if no significance). In (left) POC, \method is the only meta-learner with no difference from both EUB 
and the 4-$th$ best model. In (right) ST, \method is the only meta-learner with no statistical difference from the 58-$th$ best model.
\method is statistically better than all except iForest.}\label{table:poc_pairs}
\scalebox{0.95}{
\begin{tabular}{ll}
\centering
\begin{tabular}{ll|l}
\toprule
\textbf{Ours} & \textbf{Baseline} & \textbf{p-value} \\
\midrule
\textbf{MetaOD}        & \textbf{EUB}               & 0.0522        \\
\textbf{MetaOD}        & \textbf{4-th Best}         & 0.0929        \\
\midrule
\textbf{MetaOD}        & LOF               & 0.0013                 \\
\textbf{MetaOD}        & iForest           & 0.009                  \\
\textbf{MetaOD}        & ME                & 0.0004                 \\
\textbf{MetaOD}        & GB                & 0.0051                 \\
\textbf{MetaOD}        & ISAC              & 0.0019                 \\
\textbf{MetaOD}        & \textbf{AS}                & 0.2959        \\
\textbf{MetaOD}        & \textbf{SS}                & 0.7938        \\
\textbf{MetaOD}        & ALORS             & 0.0025                 \\
\textbf{MetaOD}        & \textbf{MetaOD\_C}               & 0.6874        \\
\textbf{MetaOD}        & \textbf{MetaOD\_F}               & 0.1165        \\
\bottomrule
\end{tabular}
     & 
\begin{tabular}{ll|l}
\toprule
\textbf{Ours} & \textbf{Baseline} & \textbf{p-value} \\
\midrule
\textbf{MetaOD}        & \textbf{58-th Best}       & 0.0517           \\
\textbf{MetaOD}        & RS              & 0.0001          \\
\midrule
\textbf{MetaOD}        & LOF               & 0.0001            \\
\textbf{MetaOD}        & \textbf{iForest}           & 0.1129            \\
\textbf{MetaOD}        & ME                & 0.0001            \\
\textbf{MetaOD}        & GB                & 0.0030            \\
\textbf{MetaOD}        & ISAC              & 0.0006             \\
\textbf{MetaOD}        & AS                & 0.0009            \\
\textbf{MetaOD}        & SS                & 0.0190           \\
\textbf{MetaOD}        & ALORS             & 0.0001            \\
\textbf{MetaOD}        & MetaOD\_C               & 0.0001            \\
\textbf{MetaOD}        & MetaOD\_F               & 0.0001            \\
\bottomrule
\end{tabular}
\end{tabular}}
\end{table}

\subsubsection{Results}
In Fig.~\ref{fig:CI_POC}, we observe that
\textbf{\method is superior to all baseline methods w.r.t. the average rank and mean average precision (MAP), and perform comparably to the Empirical Upper Bound (EUB)}.
Table \ref{table:poc_pairs} (left) shows that \method is the only meta-learner that is not significantly different from both EUB (MAP=0.2051) and the 4-$th$ best model (0.2185). 
Moreover, \method is significantly better than the baselines that do not use
meta-learning 
(LOF (0.1282), iForest (0.1536), and ME (0.1382)), and the meta-learners including GB (0.1524), ISAC (0.1658) and ALORS (0.1728). 
For the full POC evaluation, see Appendix \ref{appendix:experiment_result_POC}.

\noindent \textbf{Averaging all models (ME) does not lead to good performance} as one may expect.
As shown in Fig.~\ref{fig:CI_POC}, ME is the worst performing baseline by average rank in the POC testbed.
Using a single detector, e.g., iForest, is significantly better.
This is mainly because some models perform poorly on any given dataset, and ensembling all the models indiscriminately draws overall performance down. 
Using selective ensembles \cite{rayana2016less} could be beneficial, however, ME is too expensive to use in practice. Rather a fast, low overhead meta-learner is preferable.

\noindent 
\textbf{Meta-learners perform significantly better than methods without model selection.}
In particular, four meta-learners (\method, SS, \methodc, \methodf) significantly outperform single outlier detection methods (LOF and iForest) as well as the Mega Ensemble (ME) that averages all the models. 
\method respectively has 58.74\%, 32.48\%, and 47.25\% higher MAP over LOF, iForest, and ME. 
These results signify the benefits of model selection.

\noindent \textbf{Optimization-based meta learners generally perform better than simple meta learners}. 
The top-3 meta learners by average rank (\method, SS, and \methodc) 
are all optimization-based and significantly outperform simple meta-learners like ISAC 
as shown in Fig.~\ref{fig:CI_POC}.
Simple meta-learners weigh meta-features equally for task similarity, whereas optimization-based methods learn which meta-features matter (e.g., regression on meta-features), leading to better results.
We find that \method achieves 33.53\%, 22.74\%, and 4.73\% higher MAP than simple meta-learners including GB, ISAC, and AS. 

\subsection{ST Testbed Results} 
\label{sec:exp-st-testbed}

\subsubsection{Testbed Setting}
\label{subsec:ST}
When meta-train datasets lack similarity to the test dataset, it is hard to capitalize on prior experience.
In the extreme case, meta-learning may not perform better than no-model-selection baselines, e.g., a single detector. 
To investigate the impact of the train/test similarity on meta-learning performance, we build the ST testbed that consists of 62 datasets (Appendix \ref{appendix:databases} Table \ref{table:ST_dataset}) with relatively low similarity as shown in Fig.~\ref{fig:meta_feature}.
For evaluation on ST, we use leave-one-out cross validation; 
each time using 61 datasets as meta-train.

\begin{figure}[!tp]
\centering
    \includegraphics[scale=0.42]{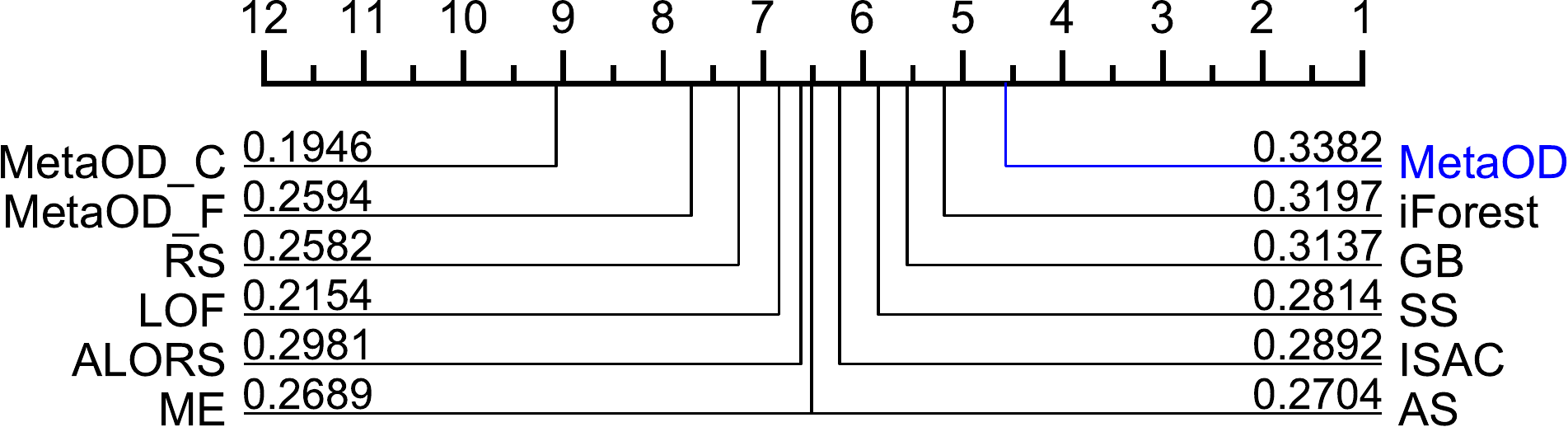}
    \vspace{-0.1in}
    \caption{Comparison of avg. performance rank (lower is better) of methods across datasets in ST. Mean AP (higher is better)  shown on lines. \method outperforms all baselines.}
\label{fig:CI_ST}
\end{figure}

\subsubsection{Results}
For the ST testbed, \textbf{\method still outperforms all baseline methods w.r.t. average rank and MAP} as shown in Fig.~\ref{fig:CI_ST}. 
Table \ref{table:poc_pairs} (right) shows that 
\method (0.3382) could select, from a pool of 302, the model that is as good as the 58-$th$ best model per dataset (0.3513). 
The comparable model changes from the 4-$th$ best per dataset in POC to 58-$th$ best in ST, which is expected due to the lower task similarity to leverage in ST. 
Notably, all other baselines are worse than the 80-$th$ best model with statistical significance. 
Moreover, \method is significantly better than all baselines except iForest. 
Note that {\method is significantly better than RS, showing that it is able to exploit the meta-train database despite limited task similarity and not simply resorting to random picking}. 
These results suggest that
\textbf{\method is a good choice under various extent of similarity among train/test datasets}.
We refer to Appendix \ref{appendix:experiment_result_ST} for detailed ST results on individual ST datasets.

\vspace{0.05in}

\noindent \textbf{Training stability affects performance for optimization-based methods.} 
Notably, several optimization-based meta-learners, such as ALORS and \methodc, do not perform well for ST. 
We find that the training process of matrix factorization is not stable when latent similarities are weak. 
In \method, we employ two strategies that help stabilize the training. 
First, we leverage meta-feature based (rather than random) initialization.
Second, we use cyclical learning rates that help escape saddle points for better local optima \cite{smith2017cyclical}.
Consequently, \method (0.3382) significantly outperforms ALORS (0.2981) and \methodc (0.1946) with 13.45\% and 73.79\% higher MAP.

\vspace{0.05in}

\noindent \textbf{Global methods outperform local methods under limited task similarity.}
In ST, datasets are less similar and simple meta-learners that leverage
task similarity locally often perform poorly.
For example, AS selects the model based on the 1NN, and is likely to fail if the most similar meta-train task is still quite dissimilar to the current task.
On the other hand, the global meta-learner GB outperforms ISAC and AS. 
Note the opposite ordering among these methods in POC as shown in Fig. \ref{fig:CI_POC}. 
As such, \textbf{effectiveness of simple meta-learners highly depends on the train/test dataset similarity}, making them harder to use in general. In contrast, \method performs well in both settings.

\vspace{-0.05in}
\subsection{Runtime Analysis}
\begin{figure}[!t]
\centering
    \includegraphics[scale=0.6]{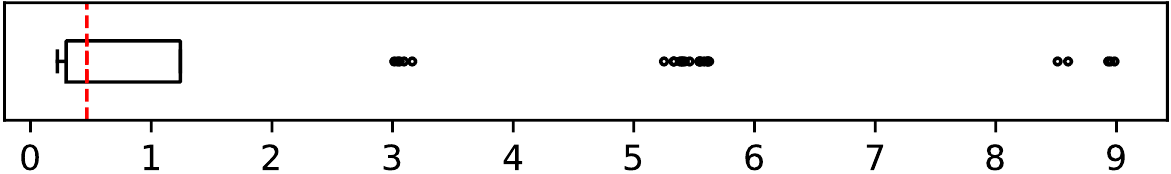}
    \vspace{-0.12in}
    \caption{\label{fig:model_select_time}
    Boxplot of \method time in POC (in sec.).  
    \method takes less than 1 sec (med. $=0.48$ shown in red) for most datasets.
    } 
    \vspace{-0.15in}
\end{figure}

\begin{figure}[!t]
\centering
    \includegraphics[scale=0.6]{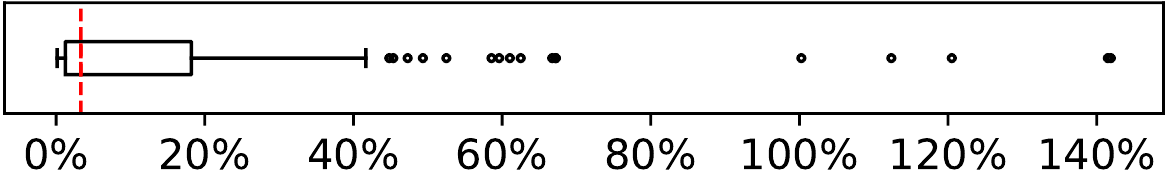}
    \vspace{-0.12in}
    \caption{\label{fig:model_select_perc}
    Boxplot of percentage of time \method takes relative to the training of the selected model in POC. \method incurs negligible overhead (med. $=3.3\%$ shown in red). 
    } 
\end{figure}

\vspace{-0.025in}
\label{sec:exp-runtime}
Fig. \ref{fig:model_select_time} shows that \method (meta-feature generation and model selection) takes less than 1 second on most datasets, while Fig. \ref{fig:model_select_perc} shows that it  incurs only negligible overhead relative to actual training of the selected model ($\approx$10\% on average). See Appendix \ref{appendix:runtime} for more comparison.

\vspace{-0.05in}
\section{Conclusion}
\label{sec:conclusion}

We introduced (to our knowledge) the {\em first systematic model selection approach to unsupervised outlier detection (OD)}. The proposed \method is a meta-learner, and builds on an extensive pool of historical outlier detection datasets and models. Given a new task, it selects a model based on the past performances of models on similar historical tasks. To effectively capture task similarity, we designed novel problem-specific meta-features.
Importantly, \method is ($i$) fully unsupervised, requiring no model evaluations at test time, and ($ii$) lightweight, incurring relatively small selection time overhead prior to model estimation. Extensive experiments on two large testbeds showed that 
\method significantly  improves detection performance over 
directly using some of the most popular models as well as several state-of-the-art unsupervised meta-learners.  

We open-source \method for future deployment.
We expect  meta-learning to get more powerful as the meta-train database 
grows.
Therefore, we also share all our code and testbeds for the community to contribute new datasets and models to stimulate further advances on automating OD. 




\clearpage
\newpage

\nocite{langley00}

\bibliography{ref}

\begin{thebibliography}{55}
\providecommand{\natexlab}[1]{#1}
\providecommand{\url}[1]{\texttt{#1}}
\expandafter\ifx\csname urlstyle\endcsname\relax
  \providecommand{\doi}[1]{doi: #1}\else
  \providecommand{\doi}{doi: \begingroup \urlstyle{rm}\Url}\fi

\bibitem[Abdulrahman et~al.(2018)Abdulrahman, Brazdil, van Rijn, and
  Vanschoren]{abdulrahman2018speeding}
Abdulrahman, S.~M., Brazdil, P., van Rijn, J.~N., and Vanschoren, J.
\newblock Speeding up algorithm selection using average ranking and active
  testing by introducing runtime.
\newblock \emph{Mach. Learn.}, 107\penalty0 (1):\penalty0 79--108, 2018.
\newblock URL
  \url{http://dblp.uni-trier.de/db/journals/ml/ml107.html#AbdulrahmanBRV18}.

\bibitem[Aggarwal(2013)]{books/sp/Aggarwal2013}
Aggarwal, C.~C.
\newblock \emph{Outlier Analysis}.
\newblock Springer, 2013.
\newblock ISBN 978-1-4614-6396-2.
\newblock URL \url{http://dx.doi.org/10.1007/978-1-4614-6396-2}.

\bibitem[Bergstra \& Bengio(2012)Bergstra and Bengio]{bergstra2012random}
Bergstra, J. and Bengio, Y.
\newblock Random search for hyper-parameter optimization.
\newblock \emph{J. Mach. Learn. Res.}, 13:\penalty0 281--305, 2012.
\newblock URL
  \url{http://dblp.uni-trier.de/db/journals/jmlr/jmlr13.html#BergstraB12}.

\bibitem[Bharadhwaj(2019)]{bharadhwaj2019meta}
Bharadhwaj, H.
\newblock Meta-learning for user cold-start recommendation.
\newblock In \emph{IJCNN}, pp.\  1--8. IEEE, 2019.
\newblock ISBN 978-1-7281-1985-4.
\newblock URL
  \url{http://dblp.uni-trier.de/db/conf/ijcnn/ijcnn2019.html#Bharadhwaj19}.

\bibitem[Breunig et~al.(2000)Breunig, Kriegel, Ng, and Sander]{breunig2000lof}
Breunig, M.~M., Kriegel, H.-P., Ng, R.~T., and Sander, J.
\newblock Lof: Identifying density-based local outliers.
\newblock In Chen, W., Naughton, J.~F., and Bernstein, P.~A. (eds.),
  \emph{SIGMOD Conference}, pp.\  93--104. ACM, 2000.
\newblock ISBN 1-58113-217-4.
\newblock URL
  \url{http://dblp.uni-trier.de/db/conf/sigmod/sigmod2000.html#BreunigKNS00}.
\newblock SIGMOD Record 29(2), June 2000.

\bibitem[Burnaev et~al.(2015)Burnaev, Erofeev, and Smolyakov]{burnaev2015model}
Burnaev, E., Erofeev, P., and Smolyakov, D.
\newblock Model selection for anomaly detection.
\newblock In Verikas, A., Radeva, P., and Nikolaev, D.~P. (eds.), \emph{ICMV},
  volume 9875 of \emph{SPIE Proceedings}, pp.\  987525. SPIE, 2015.
\newblock ISBN 9781510601161.
\newblock URL
  \url{http://dblp.uni-trier.de/db/conf/icmv/icmv2015.html#BurnaevES15}.

\bibitem[Campos et~al.(2016)Campos, Zimek, Sander, Campello, Micenková,
  Schubert, Assent, and Houle]{Campos2016}
Campos, G.~O., Zimek, A., Sander, J., Campello, R. J. G.~B., Micenková, B.,
  Schubert, E., Assent, I., and Houle, M.~E.
\newblock On the evaluation of unsupervised outlier detection: measures,
  datasets, and an empirical study.
\newblock \emph{Data Min. Knowl. Discov.}, 30\penalty0 (4):\penalty0 891--927,
  2016.
\newblock URL
  \url{http://dblp.uni-trier.de/db/journals/datamine/datamine30.html#CamposZSCMSAH16}.

\bibitem[Chen et~al.(2017)Chen, Sathe, Aggarwal, and Turaga]{chen2017outlier}
Chen, J., Sathe, S., Aggarwal, C.~C., and Turaga, D.~S.
\newblock Outlier detection with autoencoder ensembles.
\newblock In Chawla, N.~V. and Wang, W. (eds.), \emph{Proceedings of the 2017
  {SIAM} International Conference on Data Mining, Houston, Texas, USA, April
  27-29, 2017}, pp.\  90--98. {SIAM}, 2017.
\newblock \doi{10.1137/1.9781611974973.11}.
\newblock URL \url{https://doi.org/10.1137/1.9781611974973.11}.

\bibitem[Deng \& Xu(2007)Deng and Xu]{Deng07}
Deng, H. and Xu, R.
\newblock Model selection for anomaly detection in wireless ad hoc networks.
\newblock In \emph{CIDM}, pp.\  540--546. IEEE, 2007.
\newblock ISBN 1-4244-0705-2.
\newblock URL
  \url{http://dblp.uni-trier.de/db/conf/cidm/cidm2007.html#DengX07}.

\bibitem[Emmott et~al.(2016)Emmott, Das, Dietterich, Fern, and
  Wong]{Emmott2016AnomalyDM}
Emmott, A., Das, S., Dietterich, T.~G., Fern, A., and Wong, W.-K.
\newblock Anomaly detection meta-analysis benchmarks.
\newblock 2016.
\newblock URL
  \url{https://ir.library.oregonstate.edu/concern/datasets/47429f155}.

\bibitem[Evangelista et~al.(2007)Evangelista, Embrechts, and
  Szymanski]{evangelista2007some}
Evangelista, P.~F., Embrechts, M.~J., and Szymanski, B.~K.
\newblock Some properties of the gaussian kernel for one class learning.
\newblock In de~Sá, J.~M., Alexandre, L.~A., Duch, W., and Mandic, D.~P.
  (eds.), \emph{ICANN (1)}, volume 4668 of \emph{Lecture Notes in Computer
  Science}, pp.\  269--278. Springer, 2007.
\newblock ISBN 978-3-540-74689-8.
\newblock URL
  \url{http://dblp.uni-trier.de/db/conf/icann/icann2007-1.html#EvangelistaES07}.

\bibitem[Fan et~al.(2019)Fan, Yue, Sarkar, and Wang]{fan2019unified}
Fan, X., Yue, Y., Sarkar, P., and Wang, Y. X.~R.
\newblock A unified framework for tuning hyperparameters in clustering
  problems.
\newblock \emph{CoRR}, abs/1910.08018, 2019.
\newblock URL
  \url{http://dblp.uni-trier.de/db/journals/corr/corr1910.html#abs-1910-08018}.

\bibitem[Feurer \& Hutter(2019)Feurer and Hutter]{feurer2019hyperparameter}
Feurer, M. and Hutter, F.
\newblock Hyperparameter optimization.
\newblock In \emph{Automated Machine Learning}, pp.\  3--33. Springer, Cham,
  2019.
\newblock URL
  \url{https://link.springer.com/chapter/10.1007/978-3-030-05318-5_1}.

\bibitem[Feurer et~al.(2015)Feurer, Springenberg, and
  Hutter]{conf/aaai/FeurerSH15}
Feurer, M., Springenberg, J.~T., and Hutter, F.
\newblock Initializing bayesian hyperparameter optimization via meta-learning.
\newblock In Bonet, B. and Koenig, S. (eds.), \emph{Proceedings of the
  Twenty-Ninth {AAAI} Conference on Artificial Intelligence, January 25-30,
  2015, Austin, Texas, {USA}}, pp.\  1128--1135. {AAAI} Press, 2015.
\newblock URL
  \url{http://www.aaai.org/ocs/index.php/AAAI/AAAI15/paper/view/10029}.

\bibitem[Feurer et~al.(2018)Feurer, Letham, and
  Bakshy]{journals/corr/abs-1802-02219}
Feurer, M., Letham, B., and Bakshy, E.
\newblock Scalable meta-learning for bayesian optimization.
\newblock \emph{CoRR}, abs/1802.02219, 2018.
\newblock URL
  \url{http://dblp.uni-trier.de/db/journals/corr/corr1802.html#abs-1802-02219}.

\bibitem[Franceschi et~al.(2017)Franceschi, Donini, Frasconi, and
  Pontil]{franceschi2017forward}
Franceschi, L., Donini, M., Frasconi, P., and Pontil, M.
\newblock Forward and reverse gradient-based hyperparameter optimization.
\newblock In Precup, D. and Teh, Y.~W. (eds.), \emph{Proceedings of the 34th
  International Conference on Machine Learning, {ICML} 2017, Sydney, NSW,
  Australia, 6-11 August 2017}, volume~70 of \emph{Proceedings of Machine
  Learning Research}, pp.\  1165--1173. {PMLR}, 2017.
\newblock URL \url{http://proceedings.mlr.press/v70/franceschi17a.html}.

\bibitem[Fréry et~al.(2017)Fréry, Habrard, Sebban, Caelen, and
  He-Guelton]{frery2017efficient}
Fréry, J., Habrard, A., Sebban, M., Caelen, O., and He-Guelton, L.
\newblock Efficient top rank optimization with gradient boosting for supervised
  anomaly detection.
\newblock In Ceci, M., Hollmén, J., Todorovski, L., Vens, C., and Dzeroski, S.
  (eds.), \emph{ECML/PKDD (1)}, volume 10534 of \emph{Lecture Notes in Computer
  Science}, pp.\  20--35. Springer, 2017.
\newblock ISBN 978-3-319-71249-9.
\newblock URL
  \url{http://dblp.uni-trier.de/db/conf/pkdd/pkdd2017-1.html#FreryHSCH17}.

\bibitem[Goldstein \& Dengel(2012)Goldstein and Dengel]{goldstein2012histogram}
Goldstein, M. and Dengel, A.
\newblock Histogram-based outlier score (hbos): A fast unsupervised anomaly
  detection algorithm.
\newblock \emph{KI-2012: Poster and Demo Track}, pp.\  59--63, 2012.
\newblock URL
  \url{https://citeseerx.ist.psu.edu/viewdoc/download?doi=10.1.1.401.5686&rep=rep1&type=pdf}.

\bibitem[He et~al.(2021)He, Zhao, and Chu]{he2019automl}
He, X., Zhao, K., and Chu, X.
\newblock Automl: {A} survey of the state-of-the-art.
\newblock \emph{Knowl. Based Syst.}, 212:\penalty0 106622, 2021.
\newblock \doi{10.1016/j.knosys.2020.106622}.
\newblock URL \url{https://doi.org/10.1016/j.knosys.2020.106622}.

\bibitem[Hutter et~al.(2011)Hutter, Hoos, and
  Leyton-Brown]{hutter2011sequential}
Hutter, F., Hoos, H.~H., and Leyton-Brown, K.
\newblock Sequential model-based optimization for general algorithm
  configuration.
\newblock In Coello, C. A.~C. (ed.), \emph{LION}, volume 6683 of \emph{Lecture
  Notes in Computer Science}, pp.\  507--523. Springer, 2011.
\newblock ISBN 978-3-642-25565-6.
\newblock URL
  \url{http://dblp.uni-trier.de/db/conf/lion/lion2011.html#HutterHL11}.

\bibitem[Idé \& Kashima(2004)Idé and Kashima]{shyu2003novel}
Idé, T. and Kashima, H.
\newblock Eigenspace-based anomaly detection in computer systems.
\newblock In Kim, W., Kohavi, R., Gehrke, J., and DuMouchel, W. (eds.),
  \emph{KDD}, pp.\  440--449. ACM, 2004.
\newblock ISBN 1-58113-888-1.
\newblock URL \url{http://dblp.uni-trier.de/db/conf/kdd/kdd2004.html#IdeK04}.

\bibitem[Jones et~al.(1998)Jones, Schonlau, and Welch]{journals/jgo/JonesSW98}
Jones, D.~R., Schonlau, M., and Welch, W.~J.
\newblock Efficient global optimization of expensive black-box functions.
\newblock \emph{J. Global Optimization}, 13\penalty0 (4):\penalty0 455--492,
  1998.
\newblock URL
  \url{http://dblp.uni-trier.de/db/journals/jgo/jgo13.html#JonesSW98}.

\bibitem[Kadioglu et~al.(2010)Kadioglu, Malitsky, Sellmann, and
  Tierney]{conf/ecai/KadiogluMST10}
Kadioglu, S., Malitsky, Y., Sellmann, M., and Tierney, K.
\newblock Isac - instance-specific algorithm configuration.
\newblock In \emph{ECAI}, volume 215 of \emph{Frontiers in Artificial
  Intelligence and Applications}, pp.\  751--756. IOS Press, 2010.
\newblock URL
  \url{http://dblp.uni-trier.de/db/conf/ecai/ecai2010.html#KadiogluMST10}.

\bibitem[Kriegel et~al.(2008)Kriegel, Schubert, and Zimek]{kriegel2008angle}
Kriegel, H.-P., Schubert, M., and Zimek, A.
\newblock Angle-based outlier detection in high-dimensional data.
\newblock In Li, Y., Liu, B., and Sarawagi, S. (eds.), \emph{KDD}, pp.\
  444--452. ACM, 2008.
\newblock ISBN 978-1-60558-193-4.
\newblock URL
  \url{http://dblp.uni-trier.de/db/conf/kdd/kdd2008.html#KriegelSZ08}.

\bibitem[Lee et~al.(2019)Lee, Im, Jang, Cho, and Chung]{lee2019melu}
Lee, H., Im, J., Jang, S., Cho, H., and Chung, S.
\newblock Melu: Meta-learned user preference estimator for cold-start
  recommendation.
\newblock In Teredesai, A., Kumar, V., Li, Y., Rosales, R., Terzi, E., and
  Karypis, G. (eds.), \emph{KDD}, pp.\  1073--1082. ACM, 2019.
\newblock ISBN 978-1-4503-6201-6.
\newblock URL
  \url{http://dblp.uni-trier.de/db/conf/kdd/kdd2019.html#LeeIJCC19}.

\bibitem[Li et~al.(2017)Li, Jamieson, DeSalvo, Rostamizadeh, and
  Talwalkar]{li2017hyperband}
Li, L., Jamieson, K.~G., DeSalvo, G., Rostamizadeh, A., and Talwalkar, A.
\newblock Hyperband: A novel bandit-based approach to hyperparameter
  optimization.
\newblock \emph{J. Mach. Learn. Res.}, 18:\penalty0 185:1--185:52, 2017.
\newblock URL
  \url{http://dblp.uni-trier.de/db/journals/jmlr/jmlr18.html#LiJDRT17}.

\bibitem[Liu et~al.(2008)Liu, Ting, and Zhou]{liu2008isolation}
Liu, F.~T., Ting, K.~M., and Zhou, Z.-H.
\newblock Isolation forest.
\newblock In \emph{ICDM}, pp.\  413--422. IEEE Computer Society, 2008.
\newblock ISBN 978-0-7695-3502-9.
\newblock URL
  \url{http://dblp.uni-trier.de/db/conf/icdm/icdm2008.html#LiuTZ08}.

\bibitem[Misir \& Sebag(2017)Misir and Sebag]{journals/ai/MisirS17}
Misir, M. and Sebag, M.
\newblock Alors: An algorithm recommender system.
\newblock \emph{Artif. Intell.}, 244:\penalty0 291--314, 2017.
\newblock URL
  \url{http://dblp.uni-trier.de/db/journals/ai/ai244.html#MisirS17}.

\bibitem[Nikolic et~al.(2013)Nikolic, Maric, and Janicic]{nikolic2013simple}
Nikolic, M., Maric, F., and Janicic, P.
\newblock Simple algorithm portfolio for sat.
\newblock \emph{Artif. Intell. Rev.}, 40\penalty0 (4):\penalty0 457--465, 2013.
\newblock URL
  \url{http://dblp.uni-trier.de/db/journals/air/air40.html#NikolicMJ13}.

\bibitem[Pevný(2016)]{pevny2016loda}
Pevný, T.
\newblock Loda: Lightweight on-line detector of anomalies.
\newblock \emph{Mach. Learn.}, 102\penalty0 (2):\penalty0 275--304, 2016.
\newblock URL \url{http://dblp.uni-trier.de/db/journals/ml/ml102.html#Pevny16}.

\bibitem[Ramaswamy et~al.(2000)Ramaswamy, Rastogi, and
  Shim]{ramaswamy2000efficient}
Ramaswamy, S., Rastogi, R., and Shim, K.
\newblock Efficient algorithms for mining outliers from large data sets.
\newblock In Chen, W., Naughton, J.~F., and Bernstein, P.~A. (eds.),
  \emph{SIGMOD Conference}, pp.\  427--438. ACM, 2000.
\newblock ISBN 1-58113-217-4.
\newblock URL
  \url{http://dblp.uni-trier.de/db/conf/sigmod/sigmod2000.html#RamaswamyRS00}.
\newblock SIGMOD Record 29(2), June 2000.

\bibitem[Rayana \& Akoglu(2016)Rayana and Akoglu]{rayana2016less}
Rayana, S. and Akoglu, L.
\newblock Less is more: Building selective anomaly ensembles.
\newblock \emph{ACM Trans. Knowl. Discov. Data}, 10\penalty0 (4):\penalty0
  42:1--42:33, 2016.
\newblock URL
  \url{http://dblp.uni-trier.de/db/journals/tkdd/tkdd10.html#RayanaA16}.

\bibitem[Schölkopf et~al.(1999)Schölkopf, Williamson, Smola, Shawe-Taylor,
  and Platt]{scholkopf2000support}
Schölkopf, B., Williamson, R.~C., Smola, A.~J., Shawe-Taylor, J., and Platt,
  J.~C.
\newblock Support vector method for novelty detection.
\newblock In Solla, S.~A., Leen, T.~K., and Müller, K.-R. (eds.), \emph{NIPS},
  pp.\  582--588. The MIT Press, 1999.
\newblock ISBN 0-262-19450-3.
\newblock URL
  \url{http://dblp.uni-trier.de/db/conf/nips/nips1999.html#ScholkopfWSSP99}.

\bibitem[Schölkopf et~al.(2001)Schölkopf, Platt, Shawe-Taylor, Smola, and
  Williamson]{scholkopf2001estimating}
Schölkopf, B., Platt, J.~C., Shawe-Taylor, J., Smola, A.~J., and Williamson,
  R.~C.
\newblock Estimating the support of a high-dimensional distribution.
\newblock \emph{Neural Computation}, 13\penalty0 (7):\penalty0 1443--1471, July
  2001.
\newblock \doi{10.1162/089976601750264965}.
\newblock URL \url{https://doi.org/10.1162%2F089976601750264965}.

\bibitem[Shahriari et~al.(2016)Shahriari, Swersky, Wang, Adams, and
  de~Freitas]{shahriari2015taking}
Shahriari, B., Swersky, K., Wang, Z., Adams, R.~P., and de~Freitas, N.
\newblock Taking the human out of the loop: {A} review of bayesian
  optimization.
\newblock \emph{Proc. {IEEE}}, 104\penalty0 (1):\penalty0 148--175, 2016.
\newblock \doi{10.1109/JPROC.2015.2494218}.
\newblock URL \url{https://doi.org/10.1109/JPROC.2015.2494218}.

\bibitem[Shawi et~al.(2019)Shawi, Maher, and Sakr]{elshawi2019automated}
Shawi, R.~E., Maher, M., and Sakr, S.
\newblock Automated machine learning: State-of-the-art and open challenges.
\newblock \emph{CoRR}, abs/1906.02287, 2019.
\newblock URL \url{http://arxiv.org/abs/1906.02287}.

\bibitem[Smith(2017)]{smith2017cyclical}
Smith, L.~N.
\newblock Cyclical learning rates for training neural networks.
\newblock In \emph{WACV}, pp.\  464--472. IEEE Computer Society, 2017.
\newblock ISBN 978-1-5090-4822-9.
\newblock URL
  \url{http://dblp.uni-trier.de/db/conf/wacv/wacv2017.html#Smith17}.

\bibitem[Stern et~al.(2010)Stern, Samulowitz, Herbrich, Graepel, Pulina, and
  Tacchella]{stern2010collaborative}
Stern, D.~H., Samulowitz, H., Herbrich, R., Graepel, T., Pulina, L., and
  Tacchella, A.
\newblock Collaborative expert portfolio management.
\newblock In Fox, M. and Poole, D. (eds.), \emph{Proceedings of the
  Twenty-Fourth {AAAI} Conference on Artificial Intelligence, {AAAI} 2010,
  Atlanta, Georgia, USA, July 11-15, 2010}. {AAAI} Press, 2010.
\newblock URL
  \url{http://www.aaai.org/ocs/index.php/AAAI/AAAI10/paper/view/1857}.

\bibitem[Tang et~al.(2002)Tang, Chen, Fu, and Cheung]{tang2002enhancing}
Tang, J., Chen, Z., Fu, A.~W., and Cheung, D.~W.
\newblock Enhancing effectiveness of outlier detections for low density
  patterns.
\newblock In Cheng, M., Yu, P.~S., and Liu, B. (eds.), \emph{Advances in
  Knowledge Discovery and Data Mining, 6th Pacific-Asia Conference, {PAKDD}
  2002, Taipei, Taiwan, May 6-8, 2002, Proceedings}, volume 2336 of
  \emph{Lecture Notes in Computer Science}, pp.\  535--548. Springer, 2002.
\newblock \doi{10.1007/3-540-47887-6\_53}.
\newblock URL \url{https://doi.org/10.1007/3-540-47887-6\_53}.

\bibitem[Tax \& Duin(2004)Tax and Duin]{tax2004support}
Tax, D. M.~J. and Duin, R. P.~W.
\newblock Support vector data description.
\newblock \emph{Mach. Learn.}, 54\penalty0 (1):\penalty0 45--66, 2004.
\newblock \doi{10.1023/B:MACH.0000008084.60811.49}.
\newblock URL \url{https://doi.org/10.1023/B:MACH.0000008084.60811.49}.

\bibitem[Vaithyanathan \& Dom(1999)Vaithyanathan and
  Dom]{vaithyanathan2000generalized}
Vaithyanathan, S. and Dom, B.
\newblock Generalized model selection for unsupervised learning in high
  dimensions.
\newblock In Solla, S.~A., Leen, T.~K., and Müller, K.-R. (eds.), \emph{NIPS},
  pp.\  970--976. The MIT Press, 1999.
\newblock ISBN 0-262-19450-3.
\newblock URL
  \url{http://dblp.uni-trier.de/db/conf/nips/nips1999.html#VaithyanathanD99}.

\bibitem[Vanschoren(2018)]{journals/corr/abs-1810-03548}
Vanschoren, J.
\newblock Meta-learning: A survey.
\newblock \emph{CoRR}, abs/1810.03548, 2018.
\newblock URL
  \url{http://dblp.uni-trier.de/db/journals/corr/corr1810.html#abs-1810-03548}.

\bibitem[Vartak et~al.(2017)Vartak, Thiagarajan, Miranda, Bratman, and
  Larochelle]{conf/nips/VartakTMBL17}
Vartak, M., Thiagarajan, A., Miranda, C., Bratman, J., and Larochelle, H.
\newblock A meta-learning perspective on cold-start recommendations for items.
\newblock In Guyon, I., von Luxburg, U., Bengio, S., Wallach, H.~M., Fergus,
  R., Vishwanathan, S. V.~N., and Garnett, R. (eds.), \emph{Advances in Neural
  Information Processing Systems 30: Annual Conference on Neural Information
  Processing Systems 2017, December 4-9, 2017, Long Beach, CA, {USA}}, pp.\
  6904--6914, 2017.
\newblock URL
  \url{https://proceedings.neurips.cc/paper/2017/hash/51e6d6e679953c6311757004d8cbbba9-Abstract.html}.

\bibitem[Wistuba et~al.(2015)Wistuba, Schilling, and
  Schmidt-Thieme]{conf/dsaa/WistubaSS15}
Wistuba, M., Schilling, N., and Schmidt-Thieme, L.
\newblock Learning hyperparameter optimization initializations.
\newblock In \emph{DSAA}, pp.\  1--10. IEEE, 2015.
\newblock URL
  \url{http://dblp.uni-trier.de/db/conf/dsaa/dsaa2015.html#WistubaSS15}.

\bibitem[Wistuba et~al.(2018)Wistuba, Schilling, and
  Schmidt-Thieme]{journals/ml/WistubaSS18}
Wistuba, M., Schilling, N., and Schmidt-Thieme, L.
\newblock Scalable gaussian process-based transfer surrogates for
  hyperparameter optimization.
\newblock \emph{Mach. Learn.}, 107\penalty0 (1):\penalty0 43--78, 2018.
\newblock URL
  \url{http://dblp.uni-trier.de/db/journals/ml/ml107.html#WistubaSS18}.

\bibitem[Wolpert \& Macready(1997)Wolpert and
  Macready]{journals/tec/DolpertM97}
Wolpert, D.~H. and Macready, W.~G.
\newblock No free lunch theorems for optimization.
\newblock \emph{IEEE Trans. Evolutionary Computation}, 1\penalty0 (1):\penalty0
  67--82, 1997.
\newblock URL
  \url{http://dblp.uni-trier.de/db/journals/tec/tec1.html#DolpertM97}.

\bibitem[Xiao et~al.(2014)Xiao, Wang, Zhang, and Xu]{xiao2014two}
Xiao, Y., Wang, H., Zhang, L., and Xu, W.
\newblock Two methods of selecting gaussian kernel parameters for one-class svm
  and their application to fault detection.
\newblock \emph{Knowl. Based Syst.}, 59:\penalty0 75--84, 2014.
\newblock URL
  \url{http://dblp.uni-trier.de/db/journals/kbs/kbs59.html#XiaoWZX14}.

\bibitem[Xu et~al.(2012)Xu, Hutter, Shen, Hoos, and
  Leyton-Brown]{xu2012satzilla2012}
Xu, L., Hutter, F., Shen, J., Hoos, H.~H., and Leyton-Brown, K.
\newblock Satzilla2012: Improved algorithm selection based on cost-sensitive
  classification models.
\newblock \emph{Proceedings of SAT Challenge}, pp.\  57--58, 2012.
\newblock URL
  \url{http://citeseerx.ist.psu.edu/viewdoc/summary?doi=10.1.1.261.668}.

\bibitem[Yang \& Shami(2020)Yang and Shami]{yang2020hyperparameter}
Yang, L. and Shami, A.
\newblock On hyperparameter optimization of machine learning algorithms: Theory
  and practice.
\newblock \emph{Neurocomputing}, 415:\penalty0 295--316, 2020.
\newblock URL
  \url{http://dblp.uni-trier.de/db/journals/ijon/ijon415.html#YangS20}.

\bibitem[Yao et~al.(2018)Yao, Wang, Escalante, Guyon, Hu, Li, Tu, Yang, and
  Yu]{yao2018taking}
Yao, Q., Wang, M., Escalante, H.~J., Guyon, I., Hu, Y., Li, Y., Tu, W., Yang,
  Q., and Yu, Y.
\newblock Taking human out of learning applications: {A} survey on automated
  machine learning.
\newblock \emph{CoRR}, abs/1810.13306, 2018.
\newblock URL \url{http://arxiv.org/abs/1810.13306}.

\bibitem[Yu \& Zhu(2020)Yu and Zhu]{yu2020hyperparameter}
Yu, T. and Zhu, H.
\newblock Hyper-parameter optimization: {A} review of algorithms and
  applications.
\newblock \emph{CoRR}, abs/2003.05689, 2020.
\newblock URL \url{https://arxiv.org/abs/2003.05689}.

\bibitem[Zhao et~al.(2019)Zhao, Nasrullah, and Li]{zhao2019pyod}
Zhao, Y., Nasrullah, Z., and Li, Z.
\newblock Pyod: A python toolbox for scalable outlier detection.
\newblock \emph{J. Mach. Learn. Res.}, 20:\penalty0 96:1--96:7, 2019.
\newblock URL
  \url{http://dblp.uni-trier.de/db/journals/jmlr/jmlr20.html#ZhaoNL19}.

\bibitem[Zhao et~al.(2021)Zhao, Hu, Cheng, Wang, Wan, Wang, Yang, Bai, Li,
  Xiao, Wang, Qiao, Sun, and Akoglu]{zhao2021suod}
Zhao, Y., Hu, X., Cheng, C., Wang, C., Wan, C., Wang, W., Yang, J., Bai, H.,
  Li, Z., Xiao, C., Wang, Y., Qiao, Z., Sun, J., and Akoglu, L.
\newblock Suod: Accelerating large-scale unsupervised heterogeneous outlier
  detection.
\newblock \emph{Proceedings of Machine Learning and Systems}, 2021.
\newblock URL \url{https://arxiv.org/abs/2002.03222}.

\bibitem[Zheng et~al.(2007)Zheng, Yang, and Zhu]{zheng2007initialization}
Zheng, Z., Yang, J., and Zhu, Y.
\newblock Initialization enhancer for non-negative matrix factorization.
\newblock \emph{Eng. Appl. Artif. Intell.}, 20\penalty0 (1):\penalty0 101--110,
  2007.
\newblock URL
  \url{http://dblp.uni-trier.de/db/journals/eaai/eaai20.html#ZhengYZ07}.

\bibitem[Zhou \& Paffenroth(2017)Zhou and Paffenroth]{zhou2017anomaly}
Zhou, C. and Paffenroth, R.~C.
\newblock Anomaly detection with robust deep autoencoders.
\newblock In \emph{Proceedings of the 23rd {ACM} {SIGKDD} International
  Conference on Knowledge Discovery and Data Mining, Halifax, NS, Canada,
  August 13 - 17, 2017}, pp.\  665--674. {ACM}, 2017.
\newblock \doi{10.1145/3097983.3098052}.
\newblock URL \url{https://doi.org/10.1145/3097983.3098052}.

\end{thebibliography}
\bibliographystyle{icml2021}

\clearpage
\newpage

\appendix
\section*{Supplementary Material: Automating Outlier Detection via Meta-Learning}
\textit{Details on Models, Meta-features, Datasets/Testbeds, Optimization, and Detailed Experiment Result}

\section{\method Model Set}
\label{appendix:model_set}
Model set $\mM$ is composed by pairing outlier detection algorithms to distinct hyperparameter choices. 
Table \ref{table:models_long} provides a comprehensive description of models, including 302 unique models composed by 8 popular outlier detection (OD) algorithms. All models and parameters are based on the Python Outlier Detection Toolbox (PyOD)\footnote{\url{https://github.com/yzhao062/pyod}}.

\begin{table*}[!t]
	\vspace{-0.1in}
\footnotesize
\centering
	\caption{Outlier Detection Models; see hyperparameter definitions from PyOD \cite{zhao2019pyod}} 
	   \scalebox{0.8}{
	\begin{tabular}{lll|r} 
	    \toprule
		\textbf{Detection algorithm} &
		\textbf{Hyperparameter 1} & \textbf{Hyperparameter 2} & \textbf{Total}\\
		\midrule
		
		   LOF \cite{breunig2000lof}                              & n\_neighbors: $[1, 5 ,10, 15, 20, 25, 50, 60, 70, 80, 90, 100]$   & distance: ['manhattan', 'euclidean', 'minkowski'] & 36\\

		    kNN \cite{ramaswamy2000efficient}                      & n\_neighbors: $[1, 5 ,10, 15, 20, 25, 50, 60, 70, 80, 90, 100]$   & method: ['largest', 'mean', 'median'] & 36\\

   		    OCSVM \cite{scholkopf2001estimating}                   & nu (train error tol): $[0.1, 0.2, 0.3, 0.4, 0.5, 0.6, 0.7, 0.8, 0.9]$    & kernel: ['linear', 'poly', 'rbf', 'sigmoid']  & 36\\

	COF \cite{tang2002enhancing}                           & n\_neighbors: $[3, 5, 10, 15, 20, 25, 50]$   & N/A & 7\\

		  ABOD \cite{kriegel2008angle}                           & n\_neighbors: $[3, 5, 10, 15, 20, 25, 50, 60, 70, 80, 90, 100]$   & N/A  & 7\\

         iForest \cite{liu2008isolation}	    	         	& n\_estimators: $[10, 20, 30, 40, 50, 75, 100, 150, 200]$  & max\_features: $[0.1, 0.2, 0.3, 0.4, 0.5, 0.6, 0.7, 0.8, 0.9]$  & 81\\
   
         HBOS \cite{goldstein2012histogram}                     & n\_histograms: $[5, 10, 20, 30, 40, 50, 75, 100]$  & tolerance: $[0.1, 0.2, 0.3, 0.4, 0.5]$   & 40\\

		LODA \cite{pevny2016loda}                              & n\_bins: $[10, 20, 30, 40, 50, 75, 100, 150, 200]$        & n\_random\_cuts: $[5, 10, 15, 20, 25, 30]$ & 54\\

		\midrule
		&&& \textbf{302} \\ 
	\end{tabular}}
	\label{table:models_long} 
	\vspace{-0.25in}
\end{table*}

\section{OD Meta-Features}
\label{appendix:od_meta_features}

\subsection{Complete List of Features}
\label{appendix:sub_list_features}
We summarize the meta-features used by \method in Table \ref{table:meta_features}.
When applicable, we provide the formula for computing the meta-feature(s) and corresponding variants. Some are based on \cite{journals/corr/abs-1810-03548}.
Refer to the accompanied code for details. 

Specifically, meta-features can be categorized into (1) statistical features, and (2) landmarker features. 
Broadly speaking, the former captures statistical  properties of the underlying data distributions; e.g., min, max, variance, skewness, covariance, etc. of the features and  feature combinations. 
These statistics-based meta-features have been commonly used in the AutoML literature \cite{journals/corr/abs-1810-03548}.

\begin{table*}
\footnotesize
\centering
	\caption{Selected meta-features for characterizing an arbitrary dataset. See code for details.} 
	   \scalebox{0.88}{
	\begin{tabular}{l l  l l  } 
		\toprule
		Name  & Formula & Rationale & Variants \\
		\midrule
        Nr instances      &  n & Speed, Scalability         & $\frac{p}{n}$, $\log(n)$, $\log(\frac{n}{p})$ \\ 
        Nr features       &  p & Curse of dimensionality    & $\log(p)$, \% categorical \\ 
		Sample mean & $\mu$ & Concentration&  \\
		Sample median & $\Tilde{X}$ &Concentration&\\
		Sample var & $\sigma^2$ & Dispersion&\\
		Sample min &$\max_{X}$& Data range&\\
		Sample max & $\min_{X}$& Data range &\\
		Sample std & $\sigma$ & Dispersion&\\
		Percentile & $P_i$& Dispersion &q1, q25, q75, q99\\
		Interquartile Range (IQR) &$q75-q25$& Dispersion& \\
		Normalized mean & $\frac{\mu}{\max_{X}}$& Data range&\\
		Normalized median & $\frac{\Tilde{X}}{\max_{X}}$ & Data range&\\
		Sample range & $\max_{X} - \min_{X}$& Data range&\\
		Sample Gini &&Dispersion&\\
		Median absolute deviation & $\text{median}(X-\Tilde{X})$& Variability and dispersion&\\
        Average absolute deviation &$\text{avg}(X-\Tilde{X})$&Variability and dispersion&\\
        Quantile Coefficient Dispersion &$\frac{(q75 - q25)}{(q75 + q25)} $&Dispersion&\\
        Coefficient of variance  & & Dispersion&\\
        Outlier outside 1 \& 99 & \% samples outside 1\% or 99\% &Basic outlying patterns& \\
        Outlier 3 STD & \% samples outside $3\sigma$ &Basic outlying patterns& \\
		\midrule
		
		Normal test &If a sample differs from a normal dist.& Feature normality& \\
		$k$th moments &&& 5th to 10th moments\\ 
		Skewness  & Feature skewness& Feature normality & $\min$, $\max$, $\mu$, $\sigma$, $\text{skewness}$, $\text{kurtosis}$\\
        Kurtosis  & $\frac{\mu_4}{\sigma^4}$ & Feature normality & $\min$, $\max$, $\mu$, $\sigma$, $\text{skewness}$, $\text{kurtosis}$\\
        Correlation  & $\rho$& Feature interdependence & $\min$, $\max$, $\mu$, $\sigma$, $\text{skewness}$, $\text{kurtosis}$\\
        Covariance  & Cov & Feature interdependence & $\min$, $\max$, $\mu$, $\sigma$, $\text{skewness}$, $\text{kurtosis}$\\
        Sparsity & $\frac{\# \text{Unique values}}{n}$& Degree of discreteness & $\min$, $\max$, $\mu$, $\sigma$, $\text{skewness}$, $\text{kurtosis}$\\
        ANOVA p-value & $p_{\text{ANOVA}}$& Feature redundancy  & $\min$, $\max$, $\mu$, $\sigma$, $\text{skewness}$, $\text{kurtosis}$\\
        Coeff of variation & $\frac{\sigma_x}{\mu_x}$& Dispersion & \\
        Norm. entropy & $\frac{H(X)}{log_2n}$& Feature informativeness  & min,max, $\sigma$, $\mu$\\
        \midrule
        Landmarker (HBOS) & See \S \ref{appendix:landmarker} &Outlying patterns& Histogram density\\
        Landmarker (LODA) &See \S \ref{appendix:landmarker} & Outlying patterns& Histogram density\\
        Landmarker (PCA)  &See \S \ref{appendix:landmarker}&Outlying patterns& Explained variance ratio, singular values\\
        Landmarker (iForest)&See \S \ref{appendix:landmarker}&Outlying patterns& \# of leaves, tree depth, feature importance \\
        \bottomrule
	\end{tabular}}
	\label{table:meta_features} 
\end{table*}

\subsection{Landmarker Meta-Feature Generation}
\label{appendix:landmarker}

In addition to statistical meta-features, we use four OD-specific landmarker algorithms for computing OD-specific landmarker meta-features, iForest \cite{liu2008isolation}, HBOS \cite{goldstein2012histogram}, LODA \cite{pevny2016loda}, and PCA \cite{shyu2003novel} (reconstruction error as outlier score), to capture outlying characteristics of a dataset. 
To this end, we first provide a quick overview of each algorithm and then discuss how we are using them for building meta-features. The algorithms are executed with the default parameter. Refer to the attached code for details of meta-feature construction.

\textbf{Isolation Forest (iForest)} \cite{liu2008isolation} is a tree-based ensemble method. Specifically, iForest builds a collection of base trees using the subsampled unlabeled data, splitting on (randomly selected) features as nodes. iForest grows internal nodes until the terminal leaves contain only one sample or the predefined max depth is reached. Given the max depth is not set and we have multiple base trees with each leaf containing one sample only, the anomaly score of a sample is the aggregated depth the leaves the sample falls into. The key assumption is that an anomaly is more different than the normal samples, and is, therefore, easier to be ``isolated" during the node splitting. Consequently, anomalies are closer to roots with small tree depth. For iForest, we use the balance of base trees (i.e., depth of trees and number of leaves per tree) and additional information (e.g., feature importance of each base tree). It is noted that feature importance information is available for each base tree---we therefore analyze the statistic of mean and max of base tree feature importance. Specifically, the following information of base trees are used:
\begin{compactitem}
    \item \textit{Tree depth}: min, max, mean, std, skewness, and kurtosis
    \item \textit{Number of leaves}: min, max, mean, std, skewness, and kurtosis
    \item \textit{Mean of base tree feature importance}: min, max, mean, std, skewness, and kurtosis 
    \item \textit{Max of base tree feature importance}: min, max, mean, std, skewness, and kurtosis 
\end{compactitem}

\textbf{Histogram-based Outlier Scores (HBOS)} \cite{goldstein2012histogram} assumes that each dimension (feature) of the datasets is independent. It builds a histogram on each feature to calculate the density. Given there are $n$ samples and $d$ features, for each histogram from $1...d$, HBOS estimates the sample density using all $n$ samples. Intuitively, the anomaly score of sample $g$ is defined as the sum of log of inverse density. In other words, it can be considered as an aggregation of density estimation on each feature. Obviously, the samples falling in high-density areas are more likely to be normal points and vice versa. The following information is included as part of \method:
\begin{compactitem}
    \item \textit{Mean of each histogram (per feature)}: min, max, mean, std, skewness, and kurtosis
    \item \textit{Max of each histogram (per feature) }: min, max, mean, std, skewness, and kurtosis
\end{compactitem}

\textbf{Lightweight on-line detector of anomalies (LODA)} \cite{pevny2016loda} is a fast ensemble-based anomaly detection algorithm. It shares a similar idea as HBOS---``although one one-dimensional histogram is a very weak anomaly detector, their collection yields to a strong detector". Different from HBOS that simply aggregates over all independent histograms, LODA extends the histogram-based model generating $k$ random projection vectors to compress data into one-dimensional space for building histograms. Similar to HBOS, we include in the following information as part of meta-features:
\begin{compactitem}
    \item \textit{Mean of each random projection (per feature)}: min, max, mean, std, skewness, and kurtosis
    \item \textit{Max of each random projection (per feature) }: min, max, mean, std, skewness, and kurtosis
    \item \textit{Mean of each histogram (per feature)}: min, max, mean, std, skewness, and kurtosis
    \item \textit{Max of each histogram (per feature) }: min, max, mean, std, skewness, and kurtosis
\end{compactitem}

\textbf{Principal component analysis based outlier detector (PCA)} \cite{shyu2003novel} aims to quantify sample outlyingness by projecting them into lower dimensions through principal component analysis. Since the number of normal samples is much bigger than the number of outliers, the identified projection matrix is mainly suited for normal samples. Consequently, the reconstruction error of normal samples are smaller than that of outlier samples, which can be used to measure sample outlyingness. For PCA, we include the following information into meta-features:
\begin{compactitem}
    \item \textit{Explained variance ratio on the first three principal components}: The percentage of variance it captures for the top 3 principal components
    \item \textit{Singular values}: The top 3 singular values generated during SVD process
\end{compactitem}

Additionally, we also leverage the \textbf{outlier scores by OD landmarkers} after appropriate scaling, e.g., normalization/standardization.

\section{Gradient Derivations}
\label{appendix:gradient}

In this section we provide the details for the gradient derivation of \method. 
It is organized as follow. 
We first provide a quick overview of gradient derivation in classical recommender systems, and then show the derivation of the rank-based criterion used in \method.

\subsection{Background}
Given a rating matrix $\bP \in \mathbb{R}^{n \times m}$ with $n$ users rating on $m$ items, $\bP_{ij}$ denotes $i$th user's rating on the $j$th item in classical recommender system setting. For learning the latent factors in $k$ dimensions, we try to factorize $\bP$ into user matrix $\bU \in \mathbb{R} ^{n \times k}$ and the item matrix $\bV \in \mathbb{R} ^{d \times k}$ to make $\bP \approx \bU \bV^T$. 

In classical matrix factorization setting, some entries of the performance matrix $\bP$ is missing. Consequently, one may use stochastic gradient descent to minimize the mean squared error (MSE) between $\bP$ and $\bU\bV^T$ through all non-empty entries. For each rating $\bP_{ij}$, the loss $L$ is defined as:

\begin{equation}
    L_{i,j} = L (\bU_i, \bV^{T}_{j}) = (\bP_{ij} - \bU_i \bV^{T}_{j})^2 
\end{equation}

The total loss over all non-empty entries is:
\begin{equation}
    L = \sum_{i,j} (\bP_{ij} - \bU_i \bV^{T}_{j})^2 
\end{equation}

The optimization process iterates over all non-empty entries of the performance matrix $\bP$, and updates $\bU_i$ and $\bV_j$ using the learning rate $\eta$ as:
\begin{equation}
    \bU_i \leftarrow \bU_i - \eta \frac{\partial L} {\partial \bU_i}
\end{equation}

\begin{equation}
    \bV_j \leftarrow \bV_j - \eta \frac{\partial L} {\partial \bV_j}
\end{equation}

\subsection{Rank-based Criterion}
For unsupervised OD model selection, a rank-based criteria is preferable since we are concerned most with the order of model performance other than absolute value, especially the top-1 ranked model. 
We therefore choose the negative form of discounted cumulative gain (DCG) as the objective function for dataset-wise optimization, which is defined as:
\begin{equation}
\text{DCG} = \sum_r \frac{b^{rel_r}-1}{\log_2(r+1)} \label{equ:DCG_original}
\end{equation}
where 
$rel_r$ depicts the relevance of the item 
ranked at the $r$th position and $b$ is a scalar (typically set to 2). 
In our setting, we use the performance of a model to reflect its relevance to a dataset. 
As such, DCG for dataset $i$ is re-written as
 \begin{align} \label{eq:DCG_h_appendix}
 \text{DCG}_i
 &= \sum_{j=1}^m \frac{b^{{\bP_{ij}}}-1}
 {\log_2(1+\sum_{k=1}^m \mathbbm{1}{[\hP_{ij} \leq \hP_{ik}}])}
 \end{align}
where $\hP_{ij} = \langle \bU_i, \bV_j \rangle$ is the predicted performance.
Intuitively, ranking high-performing models at the top leads to higher DCG, and a larger $b$ increases the emphasis on the quality at higher rank positions. 

A challenge with DCG is that it is not differentiable as it involves ranking/sorting.
Specifically, the sum term in the denominator of Eq. (\ref{eq:DCG_h_appendix}) uses (nonsmooth) indicator functions to obtain the position of model $j$ as ranked by the estimated performances. 
We circumvent this challenge by replacing the indicator function by the (smooth) sigmoid approximation  \cite{frery2017efficient} shown in Eq. (\ref{equ:indicator_approx}).
\begin{equation} \label{equ:indicator_approx}
    \mathbbm{1}{[\hP_{ij} \leq \hP_{ik}}] \approx \frac{1}{1 + \exp (-\alpha (\hP_{ik} - \hP_{ij}))} = \sigma(\hP_{ik} - \hP_{ij})
\end{equation}

The approximated DCG criterion for the $i$th dataset is:
{
\begin{equation}\label{eq:indicator_approx_appendix}
\text{DCG}_i  \approx \text{sDCG}_i 
= \small{\sum_{j=1}^m \frac{b^{{\bP_{ij}}}-1}{\log_2(1+\sum_{k=1}^m \sigma(\hP_{ik} - \hP_{ij}))}}
\end{equation}
}

Consequently, we optimize the smoothed loss:
\begin{equation} \label{equ:loss}
    L = - \sum_i^n \text{sDCG}_i(\bP_i, \bU_i\bV^T)
\end{equation}

which can be rewritten as:
\begin{equation}
     \min_{\bU, \bV} \;\; -\sum_{i=1}^{n} \text{DCG}_i(\bP_i, \bU_i \bV^T) \;. \label{equ:objective_appendix}
\end{equation}

\subsection{Gradient Derivation for DCG based Criterion}

As we aim to maximize the total \textit{dataset-wise} DCG, we make a pass over meta-train datasets one by one at each epoch as shown in Algorithm~\ref{algo:metod}. We update $\bU_i$ and $\bV_j$ by gradient descent as shown below. It is noted that predicted performance of $j$th model on $i$th dataset is defined as the dot product of corresponding dataset and model vector: $\hP_{ij} = \bU_i \bV^T_j$. So Eq. (\ref{eq:indicator_approx_appendix}) can be rearranged as:
\small{
\begin{align} \label{eq:simplified_objective}
-\text{DCG}_i 
&= \sum_{j=1}^m \frac{b^{{\bP_{ij}}}-1}{\log_2(1+\sum_{k=1}^m \sigma(\hP_{ik} - \hP_{ij}))} \nonumber\\
&= \ln (2) \sum_{j=1}^m \frac{ b^{{\bP_{ij}}}-1}{\ln(1+\sum_{k=1}^m \sigma(\bU_i \bV^T_k - \bU_i \bV^T_j))} \nonumber\\
&= \ln (2)\sum_{j=1}^m \frac{b^{{\bP_{ij}}}-1}{\ln(1+ \sigma (0) + \sum_{k \neq j} \sigma(\bU_i \bV^T_k - \bU_i \bV^T_j))} \nonumber\\
&= \ln (2)\sum_{j=1}^m \frac{b^{{\bP_{ij}}}-1}{\ln(\frac{3}{2} + \sum_{k \neq j} \sigma(\bU_i \bV^T_k - \bU_i \bV^T_j))} \nonumber\\
&= \ln (2)\sum_{j=1}^m \frac{b^{{\bP_{ij}}}-1}{\ln(\frac{3}{2} + \sum_{k \neq j} \sigma(\bU_i \bV^T_k - \bU_i \bV^T_j))} \nonumber\\
\end{align}
}
\normalsize

We compute the gradient of $\bU_i$ and $\bV^T_j$ as the partial derivative of $-\text{DCG}_i$ as shown in Eq. (\ref{eq:simplified_objective}). 
To ease the notation, we define:
\begin{equation} \label{eq:easy_notation1}
    w^i_{jk}=\bU_i \bV^T_k - \bU_i \bV^T_j= \langle \bU_i,  (\bV_k-\bV_j) \rangle
\end{equation}

\begin{align}\label{eq:easy_notation2}
    \beta^i_j &= \frac{3}{2}+\sum_{k \neq j} \sigma(\bU_i \bV^T_k - \bU_i \bV^T_j)  \nonumber\\
              &= \frac{3}{2}+\sum_{k \neq j} \sigma (w^i_{jk})
\end{align}

By plugging Eq. (\ref{eq:easy_notation1}) and (\ref{eq:easy_notation2}) back into Eq. (\ref{eq:simplified_objective}), it is simplified into:
\begin{align} \label{eq:full_objective}
-\text{DCG}_i &= \ln (2)\sum_{j=1}^m \frac{b^{{\bP_{ij}}}-1}{\ln(\frac{3}{2} + \sum_{k \neq j} \sigma(\bU_i \bV^T_k - \bU_i \bV^T_j))} \nonumber\\
&= \ln (2)\sum_{j=1}^m \frac{b^{{\bP_{ij}}}-1}{\ln(\beta^i_j)} \nonumber\\
\end{align}

We then obtain the gradients of $\bU_i$ and $\bV^T_j$ as follows:
\small
\begin{align} \label{eq:DCG_d_Ui}
     \frac{\partial L} {\partial \bU_i} &= \frac{\partial (-\text{DCG}_i)}{\partial \bU_i}  \nonumber \\
     & = \ln (2) \frac{\partial \left(\sum_{j=1}^m \frac{b^{{\bP_{ij}}}-1}{\ln(\beta^i_j)}\right)}{\partial \bU_i} \nonumber \\
     &= \ln{(2)} \sum_{j=1}^m \left[\frac{b^{\bP_{ij}}-1}{\beta^i_{j} \ln ^2{(\beta^i_{j})}} \sum_{k \neq j} \sigma(w^i_{jk})(1-\sigma(w^i_{jk}))(\bV_k-\bV_j)\right]
\end{align}
\normalsize

\small
\begin{align} \label{eq:DCG_d_Vj}
     \frac{\partial L} {\partial \bV_j} &= \frac{\partial (-\text{DCG}_i)}{\partial \bV_j}  \nonumber \\
     & = \ln (2) \frac{\partial \left(\sum_{j=1}^m \frac{b^{{\bP_{ij}}}-1}{\ln(\beta^i_j)}\right)}{\partial \bU_i} \nonumber \\
    &= -\ln{(2)} \sum_{j=1}^m \left[\frac{b^{\bP_{ij}}-1}{\beta^i_{j} \ln ^2{(\beta^i_{j})} } {\sum_{k \neq j} \sigma(w^i_{jk})(1-\sigma(w^i_{jk}))\bU_i}\right]
\end{align}
\normalsize

\section{\method Flowchart}
\label{sec:flow}

\begin{figure}[!ht]
\centering
    \includegraphics[scale=0.4]{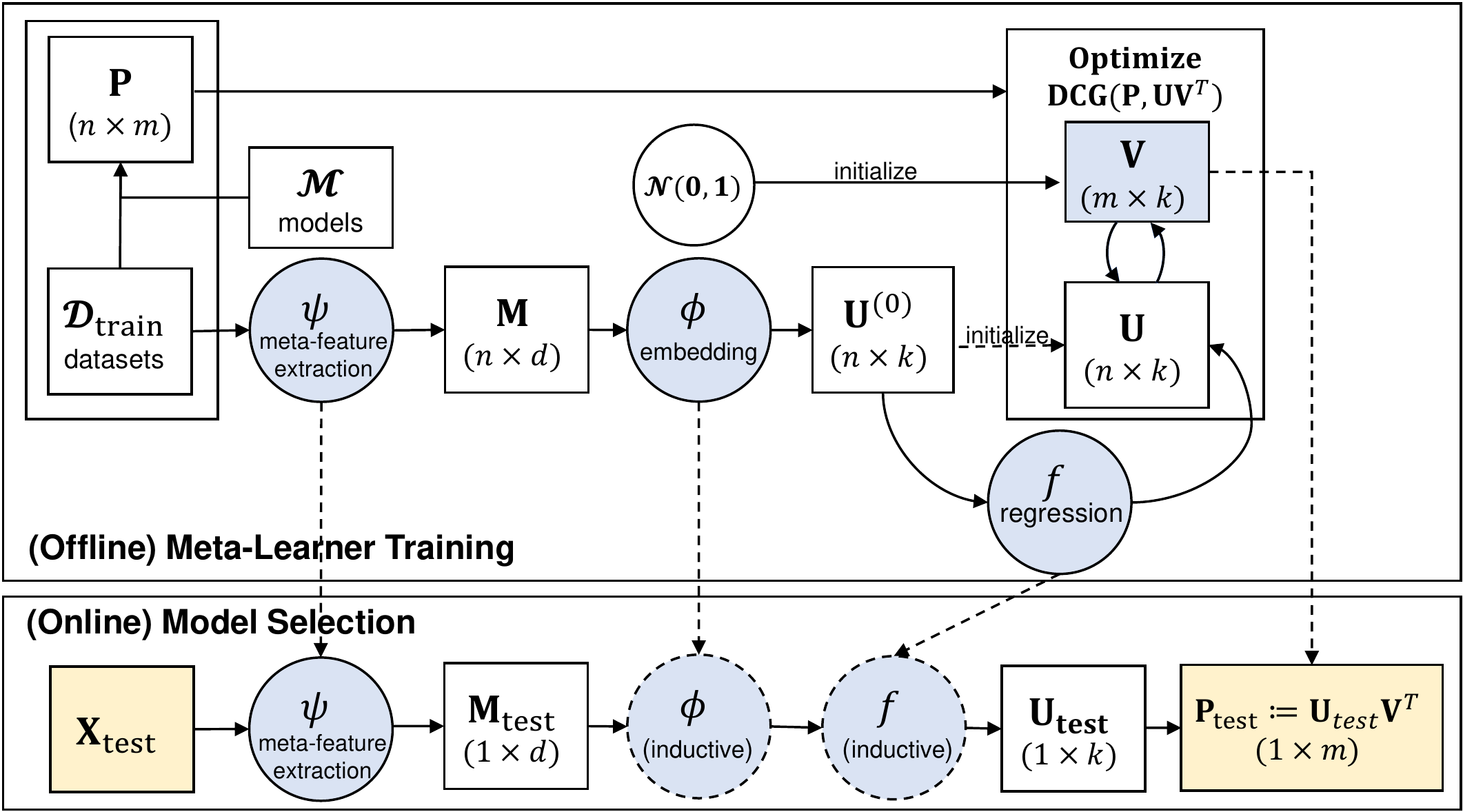} 
\vspace{-0.2in}
\caption{\label{fig:flowchart} \method overview; components that transfer from offline to online (model selection) phase shown in blue. \\}
\end{figure}

Fig. \ref{fig:flowchart} shows the major components of \method. 
We highlight the components transferred from offline to online stage (model selection) in blue; namely, meta-feature extractors ($\psi$), embedding model ($\phi$), regressor $f$, model matrix $\bU$, and dataset matrix $\bV$.

\section{Dataset Description and Testbed Setup}
\label{appendix:databases}

\subsection{POC Testbed Setup}
POC testbed is built to simulate the testbed when meta-train and test datasets come from similar distribution. Model selection on test data can therefore benefit from the prior experience on the train set. For this purpose, we use the benchmark datasets\footnote{\url{https://ir.library.oregonstate.edu/concern/datasets/47429f155}} \cite{Emmott2016AnomalyDM}. In short, they adapt 19 datasets from UCI repository and also create a synthetic dataset to make a pool of 20 ``mothersets". For each motherset, they first separate anomalies from normal points, and then generate ``childsets" from the motherset by sampling and controlling outlying properties: (i) point difficulty; (ii) relative frequency, i.e., the number of anomalies; (iii) clusteredness and (iv) feature irrelevance. Taking this approach, the childsets generated from the same motherset with the same properties are deemed to be ``siblings" with high similarity. Refer to the original paper for details of the data generation process.

We build the POC testbed by selecting five siblings from each motherset, resulting in 100 datasets. For robustness, we split the 100 datasets into 5 folds for cross-validation. Each fold contains 20 independent datasets with no siblings, and the corresponding train set (80 datasets) contain their siblings. Refer to the code for the 100 randomly selected childsets for POC testbed.

\subsection{ST Testbed Setup}
Different from the setting of POC, ST testbed aims to test out \method's performance when the meta-learning assumption does not hold: train and test datasets are all independent with limited similarity. 

To build the ST testbed, we combine the datasets from three different sources resulting in 62 independent datasets as shown in Table \ref{table:ST_dataset}: 
(i) 23 datasets from ODDS Library\footnote{\url{http://odds.cs.stonybrook.edu}} ; (ii) 19 datasets DAMI datasets \cite{Campos2016}\footnote{\url{http://www.dbs.ifi.lmu.de/research/outlier-evaluation/DAMI}} as well as (iii) 20 benchmark datasets \cite{Emmott2016AnomalyDM} used in POC. For ST testbed, we run leave-one-out cross validation. So each time 61 datasets are used for meta-train and the remaining one for test. 

\begin{table}[!htp]
\centering
	\caption{ST testbed composed by ODDS library (23 datasets), DAMI library (19 datasets), and Emmott benchmark (20 datasets)} 
	\scriptsize
	\begin{tabular}{l l | r  r r } 
	\toprule
	 & \textbf{Data}                 & \textbf{Pts} & \textbf{Dim} & \textbf{\% Outlier} \\
	\midrule
    1 & annthyroid (ODDS)             & 7200         & 6            & 7.4167                       \\
    2 & arrhythmia (ODDS)             & 452          & 274          & 14.6018                      \\
    3 & breastw (ODDS)                & 683          & 9            & 34.9927                      \\
    4 & glass (ODDS)                  & 214          & 9            & 4.2056                       \\
    5 & ionosphere (ODDS)             & 351          & 33           & 35.8974                      \\
    6 & letter (ODDS)                 & 1600         & 32           & 6.25                         \\
    7 & lympho (ODDS)                 & 148          & 18           & 4.0541                       \\
    8 & mammography (ODDS)            & 11183        & 6            & 2.325                        \\
    9 & mnist (ODDS)                  & 7603         & 100          & 9.2069                       \\
    10 & musk (ODDS)                   & 3062         & 166          & 3.1679                       \\
    11 & optdigits (ODDS)              & 5216         & 64           & 2.8758                       \\
    12 & pendigits (ODDS)              & 6870         & 16           & 2.2707                       \\
    13 & pima (ODDS)                   & 768          & 8            & 34.8958                      \\
    14 & satellite (ODDS)              & 6435         & 36           & 31.6395                      \\
    15 & satimage-2 (ODDS)             & 5803         & 36           & 1.2235                       \\
    16 & shuttle (ODDS)                & 49097        & 9            & 7.1511                       \\
    17 & smtp\_n (ODDS)                & 95156        & 3            & 0.0315                       \\
    18 & speech (ODDS)                 & 3686         & 400          & 1.6549                       \\
    19 & thyroid (ODDS)                & 3772         & 6            & 2.4655                       \\
    20 & vertebral (ODDS)              & 240          & 6            & 12.5                         \\
    21 & vowels (ODDS)                 & 1456         & 12           & 3.4341                       \\
    22 & wbc (ODDS)                    & 378          & 30           & 5.5556                       \\
    23 & wine (ODDS)                   & 129          & 13           & 7.7519                       \\
    \midrule
    24 & Annthyroid (DAMI)             & 7129         & 21           & 7.4905                       \\
    25 & Arrhythmia (DAMI)             & 450          & 259          & 45.7778                      \\
    26 & Cardiotocography (DAMI)       & 2114         & 21           & 22.0435                      \\
    27 & HeartDisease (DAMI)           & 270          & 13           & 44.4444                      \\
    28 & Hepatitis (DAMI)              & 80           & 19           & 16.25                        \\
    29 & InternetAds (DAMI)            & 1966         & 1555         & 18.7182                      \\
    30 & PageBlocks (DAMI)             & 5393         & 10           & 9.4567                       \\
    31 & Pima (DAMI)                   & 768          & 8            & 34.8958                      \\
    32 & SpamBase (DAMI)               & 4207         & 57           & 39.9097                      \\
    33 & Stamps (DAMI)                 & 340          & 9            & 9.1176                       \\
    34 & Wilt (DAMI)                   & 4819         & 5            & 5.3331                       \\
    35 & ALOI (DAMI)                   & 49534        & 27           & 3.0444                       \\
    36 & Glass (DAMI)                  & 214          & 7            & 4.2056                       \\
    37 & PenDigits (DAMI)              & 9868         & 16           & 0.2027                       \\
    38 & Shuttle (DAMI)                & 1013         & 9            & 1.2833                       \\
    39 & Waveform (DAMI)               & 3443         & 21           & 2.9044                       \\
    40 & WBC (DAMI)                    & 223          & 9            & 4.4843                       \\
    41 & WDBC (DAMI)                   & 367          & 30           & 2.7248                       \\
    42 & WPBC (DAMI)                   & 198          & 33           & 23.7374                      \\
    \midrule
    43 & abalone\_1231 (Emmott)        & 1986         & 15           & 5.0352                       \\
    44 & comm.and.crime\_0936 (Emmott) & 910          & 404          & 1.0989                       \\
    45 & concrete\_1096 (Emmott)       & 468          & 32           & 1.0684                       \\
    46 & fault\_0246 (Emmott)          & 278          & 38           & 17.9856                      \\
    47 & gas\_0321 (Emmott)            & 6000         & 128          & 0.1                          \\
    48 & imgseg\_1526 (Emmott)         & 1320         & 25           & 10                           \\
    49 & landsat\_1761 (Emmott)        & 230          & 36           & 10                           \\
    50 & letter.rec\_1666 (Emmott)     & 4089         & 23           & 10.0024                      \\
    51 & magic.gamma\_1411 (Emmott)    & 6000         & 22           & 5                            \\
    52 & opt.digits\_1316 (Emmott)     & 3180         & 248          & 5                            \\
    53 & pageb\_0126 (Emmott)          & 733          & 14           & 16.2347                      \\
    54 & particle\_1336 (Emmott)       & 6000         & 200          & 5                            \\
    55 & shuttle\_0071 (Emmott)        & 6000         & 20           & 16.3167                      \\
    56 & skin\_1706 (Emmott)           & 6000         & 4            & 10                           \\
    57 & spambase\_0681 (Emmott)       & 2522         & 57           & 0.5155                       \\
    58 & synthetic\_1786 (Emmott)      & 329          & 14           & 10.0304                      \\
    59 & wave\_0661 (Emmott)           & 3024         & 21           & 0.5291                       \\
    60 & wine\_0611 (Emmott)           & 3720         & 24           & 0.5108                       \\
    61 & yeast\_1221 (Emmott)          & 926          & 8            & 5.0756                       \\
    62 & yearp\_0231 (Emmott)          & 6000         & 202          & 48.6                        \\
    \bottomrule
	\end{tabular}
	\label{table:ST_dataset} 
\end{table}

\section{Additional Experiment Results}
\label{appendix:experiment_results}

\subsection{Experiment Results for POC Testbed}
\label{appendix:experiment_result_POC}
We present the performances of compared methods in Table \ref{table:poc_result}, and hypothesis test results in Table \ref{table:poc_pairs_full}. It is noted these results are averaged across five folds. The results shows that \method achieves the best MAP among all meta-learners.
\begin{table*}[!htp]
    \centering
    \normalsize
    \caption{Method evaluation in POC testbed (average precision). The most performing method is highlighted in \textbf{bold}. The rank is provided in parenthesis (lower ranks denote better performance). \method achieves the best average precision and average rank among all meta-learners.}\label{table:poc_result}. 
    \tiny
   \scalebox{0.85}{
    \begin{tabular}{l|llllllllll|ll}
    \toprule
\textbf{Dataset}        & \textbf{LOF} & \textbf{iForest} & \textbf{ME}     & \textbf{GB} & \textbf{ISAC}  & \textbf{AS}     & \textbf{SS}     & \textbf{ALORS}  & \textbf{MetaOD\_C} & \textbf{MetaOD\_F} & \textbf{MetaOD} & \textbf{EUB} \\
\midrule
\textbf{abalone}        & 0.0812 (12)  & 0.1679 (10)      & 0.1441 (11)         & 0.1738 (9)    & 0.192 (6)          & 0.229 (3)           & 0.224 (4)           & 0.1747 (8)     & 0.1815 (7)          & 0.2141 (5)          & 0.2329 (2)          & \textbf{0.2394 (1)}   \\
\textbf{comm.and.crime} & 0.0839 (10)  & 0.0913 (8)       & 0.0797 (11)         & 0.0855 (9)    & 0.0638 (12)        & 0.1001 (5)          & 0.1122 (2)          & 0.099 (7)      & 0.1079 (3)          & 0.0999 (6)          & 0.1072 (4)          & \textbf{0.1156 (1)}   \\
\textbf{concrete}       & 0.0297 (2)   & 0.0279 (3)       & \textbf{0.0298 (1)} & 0.0251 (5)    & 0.0224 (7)         & 0.022 (8)           & 0.0279 (3)          & 0.0236 (6)     & 0.0131 (12)         & 0.0198 (9)          & 0.0185 (10)         & 0.0147 (11)           \\
\textbf{fault}          & 0.1898 (12)  & 0.3755 (7)       & 0.323 (10)          & 0.3735 (8)    & 0.197 (11)         & 0.3949 (2)          & 0.3778 (5)          & 0.3723 (9)     & \textbf{0.4217 (1)} & 0.3813 (4)          & 0.3768 (6)          & 0.3899 (3)            \\
\textbf{gas}            & 0.0193 (5)   & 0.0031 (11)      & 0.0083 (8)          & 0.0033 (10)   & 0.0059 (9)         & \textbf{0.0481 (1)} & 0.0152 (6)          & 0.0024 (12)    & 0.0392 (2)          & 0.013 (7)           & 0.0229 (4)          & 0.0387 (3)            \\
\textbf{imgseg}         & 0.1153 (12)  & 0.3618 (6)       & 0.2586 (11)         & 0.3598 (9)    & 0.3659 (5)         & 0.3514 (10)         & 0.3891 (4)          & 0.3605 (8)     & 0.3612 (7)          & 0.3989 (3)          & 0.408 (2)           & \textbf{0.4166 (1)}   \\
\textbf{landsat}        & 0.1644 (4)   & 0.1306 (9)       & 0.131 (8)           & 0.13 (10)     & 0.1071 (12)        & 0.1578 (5)          & 0.138 (7)           & 0.1111 (11)    & \textbf{0.1844 (1)} & 0.1562 (6)          & 0.1784 (3)          & \textbf{0.1844 (1)}   \\
\textbf{letter.rec}     & 0.1529 (7)   & 0.0986 (11)      & 0.112 (9)           & 0.0991 (10)   & 0.0946 (12)        & 0.222 (2)           & 0.218 (5)           & 0.1168 (8)     & 0.2216 (3)          & \textbf{0.2229 (1)} & 0.2179 (6)          & 0.2216 (3)            \\
\textbf{magic.gamma}    & 0.1152 (9)   & 0.1303 (4)       & 0.1104 (10)         & 0.1314 (3)    & 0.1096 (11)        & 0.1319 (2)          & 0.1299 (5)          & 0.1294 (6)     & 0.102 (12)          & 0.1255 (8)          & 0.1263 (7)          & \textbf{0.1351 (1)}   \\
\textbf{opt.digits}     & 0.0662 (9)   & 0.0668 (7)       & 0.0603 (12)         & 0.0665 (8)    & 0.0742 (3)         & 0.0795 (2)          & 0.0689 (6)          & 0.0606 (11)    & 0.0705 (4)          & 0.066 (10)          & 0.0701 (5)          & \textbf{0.0803 (1)}   \\
\textbf{pageb}          & 0.3956 (11)  & 0.4581 (6)       & 0.3801 (12)         & 0.4574 (7)    & 0.4384 (9)         & 0.4829 (3)          & 0.4616 (5)          & 0.4621 (4)     & 0.4281 (10)         & 0.4498 (8)          & 0.4898 (2)          & \textbf{0.4939 (1)}   \\
\textbf{particle}       & 0.0546 (12)  & 0.0782 (5)       & 0.0626 (11)         & 0.0746 (8)    & 0.0761 (6)         & \textbf{0.1 (1)}    & 0.0936 (3)          & 0.0739 (9)     & 0.0683 (10)         & 0.0867 (4)          & 0.0757 (7)          & 0.0982 (2)            \\
\textbf{shuttle}        & 0.2015 (11)  & 0.2058 (9)       & 0.1935 (12)         & 0.2056 (10)   & 0.2961 (5)         & 0.3165 (3)          & 0.2711 (7)          & 0.2105 (8)     & 0.3165 (3)          & 0.2932 (6)          & \textbf{0.3225 (1)} & 0.3185 (2)            \\
\textbf{skin}           & 0.1161 (9)   & 0.0926 (11)      & 0.0995 (10)         & 0.0911 (12)   & 0.1814 (6)         & 0.165 (8)           & 0.2408 (4)          & 0.1737 (7)     & 0.2808 (1)          & 0.2808 (1)          & \textbf{0.2808 (1)} & 0.2278 (5)            \\
\textbf{spambase}       & 0.0187 (12)  & 0.0713 (9)       & 0.0571 (11)         & 0.0706 (10)   & 0.0873 (6)         & 0.0744 (8)          & \textbf{0.1292 (1)} & 0.0757 (7)     & 0.0981 (4)          & 0.112 (2)           & 0.0942 (5)          & 0.0982 (3)            \\
\textbf{synthetic}      & 0.1233 (4)   & 0.1226 (5)       & 0.1157 (8)          & 0.1132 (11)   & \textbf{0.154 (1)} & 0.1218 (6)          & 0.1147 (10)         & 0.1151 (9)     & 0.1468 (3)          & 0.1182 (7)          & 0.1483 (2)          & 0.1046 (12)           \\
\textbf{wave}           & 0.0577 (9)   & 0.0114 (12)      & 0.0297 (10)         & 0.0117 (11)   & 0.2925 (8)         & 0.3413 (5)          & \textbf{0.3486 (1)} & 0.3244 (7)     & \textbf{0.3486 (1)} & 0.344 (4)           & 0.3365 (6)          & \textbf{0.3486 (1)}   \\
\textbf{wine}           & 0.0082 (11)  & 0.0087 (5)       & 0.0084 (10)         & 0.0085 (8)    & 0.0085 (8)         & 0.0073 (12)         & 0.0087 (5)          & 0.0124 (2)     & 0.0097 (3)          & 0.0086 (7)          & 0.0088 (4)          & \textbf{0.0129 (1)}   \\
\textbf{yeast}          & 0.0813 (2)   & 0.0781 (4)       & 0.073 (8)           & 0.0762 (6)    & 0.0796 (3)         & 0.068 (11)          & 0.0781 (4)          & 0.0693 (9)     & \textbf{0.0885 (1)} & 0.067 (12)          & 0.0733 (7)          & 0.0688 (10)           \\
\textbf{yearp}          & 0.4894 (5)   & 0.4911 (4)       & 0.4862 (7)          & 0.4913 (3)    & 0.4703 (12)        & 0.4716 (10)         & 0.4916 (2)          & 0.4891 (6)     & 0.4716 (10)         & 0.4741 (9)          & 0.4801 (8)          & \textbf{0.4937 (1)}   \\
\midrule
\textbf{average}        & 0.1282 (8.7) & 0.1536 (7.4)     & 0.1382 (9.7) & 0.1524 (8.35) & 0.1658 (7.6)  & 0.1943 (5.49) & 0.197 (4.95) & 0.1728 (7.6)   & 0.198 (4.9)        & 0.1966 (5.95)      & 0.2035 (4.48)   & \textbf{0.2051 (3.2)}\\

\textbf{STD}            & 0.1219       & 0.1487           & 0.1294          & 0.1489      & 0.1386         & 0.1491          & 0.1486          & 0.1487          & 0.1485             & 0.1530             & 0.1587          & 0.156      \\
    \bottomrule
    \end{tabular}}
    \normalsize
    \vspace{0.05in}
\end{table*}

\begin{table*}[!h]
\scriptsize
\caption{
Pairwise statistical test results between \method and baselines by Wilcoxon signed rank test in POC. Statistically better method shown in \textbf{bold} (both marked \textbf{bold} if no significance). \method related pairs are surrounded by rectangles. \method (MAP=0.2035) is statistically significantly better than baselines including LOF (0.1282), iForest (0.1536), ME (0.1382), GB (0.1524), ISAC (0.1658), and ALORS (0.1728), and comparable to EUB (0.2051), the empirical upper bound. 
}\label{table:poc_pairs_full}
\scalebox{0.92}{
\begin{tabular}{lll}
\centering
\begin{tabular}{ll|l}
\toprule
\textbf{Method 1}         & \textbf{Method 2}           & \textbf{p-value} \\
\midrule
\textbf{LOF (0.1282)}       & \textbf{IForest (0.1536)}   & 0.3135           \\
\textbf{LOF (0.1282)}       & \textbf{ME (0.1382)}        & 0.433            \\
\textbf{LOF (0.1282)}       & \textbf{GB (0.1524)}        & 0.4553           \\
\textbf{LOF (0.1282)}       & \textbf{ISAC (0.1658)}      & 0.1005           \\
LOF (0.1282)                & \textbf{AS (0.1943)}        & 0.0025           \\
LOF (0.1282)                & \textbf{SS (0.197)}         & 0.0045           \\
\textbf{LOF (0.1282)}       & \textbf{ALORS (0.1728)}     & 0.062            \\
LOF (0.1282)                & \textbf{MetaOD\_C (0.198)}  & 0.001            \\
LOF (0.1282)                & \textbf{MetaOD\_F (0.1966)} & 0.01             \\
\marktopleft{c1}LOF (0.1282)                & \textbf{MetaOD (0.2035)} \markbottomright{c1}   & 0.0013           \\
LOF (0.1282)                & \textbf{EUB (0.2051)}       & 0.0007           \\
\textbf{IForest (0.1536)}   & ME (0.1382)                 & 0.029            \\
\textbf{IForest (0.1536)}   & GB (0.1524)                 & 0.0365           \\
\textbf{IForest (0.1536)}   & \textbf{ISAC (0.1658)}      & 0.7369           \\
IForest (0.1536)            & \textbf{AS (0.1943)}        & 0.008            \\
IForest (0.1536)            & \textbf{SS (0.197)}         & 0.0012           \\
\textbf{IForest (0.1536)}   & \textbf{ALORS (0.1728)}     & 0.6274           \\
IForest (0.1536)            & \textbf{MetaOD\_C (0.198)}  & 0.0276           \\
IForest (0.1536)            & \textbf{MetaOD\_F (0.1966)} & 0.0276           \\
\marktopleft{c2}IForest (0.1536)            & \textbf{MetaOD (0.2035)} \markbottomright{c2}   & 0.009            \\
IForest (0.1536)            & \textbf{EUB (0.2051)}       & 0.001            \\
\textbf{ME (0.1382)}        & \textbf{GB (0.1524)}        & 0.0569           \\
\bottomrule
\end{tabular}
     & 
\begin{tabular}{ll|l}
\toprule
\textbf{Method 1}         & \textbf{Method 2}           & \textbf{p-value} \\
\midrule
\textbf{ME (0.1382)}        & \textbf{ISAC (0.1658)}      & 0.156            \\
ME (0.1382)                 & \textbf{AS (0.1943)}        & 0.0005           \\
ME (0.1382)                 & \textbf{SS (0.197)}         & 0.0002           \\
ME (0.1382)                 & \textbf{ALORS (0.1728)}     & 0.009            \\
ME (0.1382)                 & \textbf{MetaOD\_C (0.198)}  & 0.0008           \\
ME (0.1382)                 & \textbf{MetaOD\_F (0.1966)} & 0.0012           \\
\marktopleft{c3}ME (0.1382)                 & \textbf{MetaOD (0.2035)} \markbottomright{c3}    & 0.0004           \\
ME (0.1382)                 & \textbf{EUB (0.2051)}       & 0.0004           \\
\textbf{GB (0.1524)}        & \textbf{ISAC (0.1658)}      & 0.7172           \\
GB (0.1524)                 & \textbf{AS (0.1943)}        & 0.0032           \\
GB (0.1524)                 & \textbf{SS (0.197)}         & 0.0002           \\
\textbf{GB (0.1524)}        & \textbf{ALORS (0.1728)}     & 0.3703           \\
GB (0.1524)                 & \textbf{MetaOD\_C (0.198)}  & 0.0169           \\
GB (0.1524)                 & \textbf{MetaOD\_F (0.1966)} & 0.0111           \\
\marktopleft{c4}GB (0.1524)                 & \textbf{MetaOD (0.2035)}\markbottomright{c4}    & 0.0051           \\
GB (0.1524)                 & \textbf{EUB (0.2051)}       & 0.0008           \\
ISAC (0.1658)               & \textbf{AS (0.1943)}        & 0.0169           \\
ISAC (0.1658)               & \textbf{SS (0.197)}         & 0.0051           \\
\textbf{ISAC (0.1658)}      & \textbf{ALORS (0.1728)}     & 0.6542           \\
\textbf{ISAC (0.1658)}      & \textbf{MetaOD\_C (0.198)}  & 0.062            \\
ISAC (0.1658)               & \textbf{MetaOD\_F (0.1966)} & 0.008            \\
\marktopleft{c5}ISAC (0.1658)               & \textbf{MetaOD (0.2035)}\markbottomright{c5}    & 0.0019           \\
\bottomrule
\end{tabular}
&
\begin{tabular}{ll|l}
\toprule
\textbf{Method 1}         & \textbf{Method 2}           & \textbf{p-value} \\
\midrule
ISAC (0.1658)               & \textbf{EUB (0.2051)}       & 0.0015           \\
\textbf{AS (0.1943)}        & \textbf{SS (0.197)}         & 0.9702           \\
\textbf{AS (0.1943)}        & ALORS (0.1728)              & 0.0206           \\
\textbf{AS (0.1943)}        & \textbf{MetaOD\_C (0.198)}  & 0.8446           \\
\textbf{AS (0.1943)}        & \textbf{MetaOD\_F (0.1966)} & 0.3135           \\
\marktopleft{c6}\textbf{AS (0.1943)}        & \textbf{MetaOD (0.2035)}\markbottomright{c6}    & 0.2959           \\
AS (0.1943)                 & \textbf{EUB (0.2051)}       & 0.0304           \\
\textbf{SS (0.197)}         & ALORS (0.1728)              & 0.0003           \\
\textbf{SS (0.197)}         & \textbf{MetaOD\_C (0.198)}  & 0.8092           \\
\textbf{SS (0.197)}         & \textbf{MetaOD\_F (0.1966)} & 0.4553           \\
\marktopleft{c7}\textbf{SS (0.197)}         & \textbf{MetaOD (0.2035)}\markbottomright{c7}    & 0.7938           \\
\textbf{SS (0.197)}         & \textbf{EUB (0.2051)}       & 0.099            \\
ALORS (0.1728)              & \textbf{MetaOD\_C (0.198)}  & 0.0276           \\
ALORS (0.1728)              & \textbf{MetaOD\_F (0.1966)} & 0.0137           \\
\marktopleft{c8}ALORS (0.1728)              & \textbf{MetaOD (0.2035)}\markbottomright{c8}    & 0.0025           \\
ALORS (0.1728)              & \textbf{EUB (0.2051)}       & 0.0006           \\
\textbf{MetaOD\_C (0.198)}  & \textbf{MetaOD\_F (0.1966)} & 0.7475           \\
\marktopleft{c9}\textbf{MetaOD\_C (0.198)}  & \textbf{MetaOD (0.2035)}\markbottomright{c9}    & 0.6874           \\
\textbf{MetaOD\_C (0.198)}  & \textbf{EUB (0.2051)}       & 0.1773           \\
\marktopleft{c10}\textbf{MetaOD\_F (0.1966)} & \textbf{MetaOD (0.2035)}\markbottomright{c10}    & 0.1165           \\
MetaOD\_F (0.1966)          & \textbf{EUB (0.2051)}       & 0.0251           \\
\marktopleft{c11}\textbf{MetaOD (0.2035)}    & \textbf{EUB (0.2051)}\markbottomright{c11}       & 0.0522           \\ 
\bottomrule
\end{tabular}
\end{tabular}}
\vspace{0.05in}
\end{table*}

\subsection{Experiment Results for ST Testbed}
\label{appendix:experiment_result_ST}
We present the method performance in Table \ref{table:st_result}, and hypothesis test result in Table \ref{table:st_pairs_full}. Among all meta-learners, \method shows the best MAP.
\begin{table*}[!htp]
    \centering
    \normalsize
    \caption{Method evaluation in ST testbed (average precision). The most performing method is highlighted in \textbf{bold}. The rank is provided in parenthesis (lower ranks denote better performance). \method achieves the best average precision and average rank among all meta-learners.}\label{table:st_result}
    \tiny
       \scalebox{0.85}{
    \begin{tabular}{l|lllllllllll|l}
    \toprule
\textbf{Datasets}         & \textbf{LOF}    & \textbf{iForest} & \textbf{ME}     & \textbf{GB}     & \textbf{ISAC}   & \textbf{AS}     & \textbf{SS}     & \textbf{ALORS}  & \textbf{MetaOD\_c} & \textbf{MetaOD\_F} & \textbf{RS}     & \textbf{MetaOD} \\
\midrule
\textbf{abalone}          & 0.092 (10)          & 0.1654 (2)          & 0.1338 (6)          & 0.1584 (3)          & 0.15 (4)            & 0.1232 (9)          & \textbf{0.1688 (1)} & 0.1316 (7)          & 0.0737 (12)         & 0.1274 (8)          & 0.092 (10)          & 0.1355 (5)          \\
\textbf{ALOI}             & 0.1424 (2)          & 0.0333 (6)          & 0.0284 (11)         & 0.0333 (6)          & 0.0329 (9)          & \textbf{0.5714 (1)} & 0.042 (4)           & 0.0282 (12)         & 0.0335 (5)          & 0.0297 (10)         & 0.0491 (3)          & 0.0333 (6)          \\
\textbf{annthyroid}       & 0.1522 (11)         & 0.2828 (7)          & 0.3177 (6)          & 0.3399 (5)          & 0.397 (2)           & \textbf{0.8089 (1)} & 0.3624 (4)          & 0.2716 (8)          & 0.0605 (12)         & 0.196 (10)          & 0.2384 (9)          & 0.3724 (3)          \\
\textbf{Annthyroid2}      & 0.1351 (2)          & 0.1198 (5)          & 0.1173 (6)          & 0.0998 (8)          & \textbf{0.1353 (1)} & 0.109 (7)           & 0.0998 (8)          & 0.127 (4)           & 0.0837 (11)         & 0.0715 (12)         & 0.1283 (3)          & 0.0998 (8)          \\
\textbf{Arrhythmia}       & 0.7435 (5)          & 0.7615 (2)          & 0.6643 (9)          & \textbf{0.7622 (1)} & 0.7471 (4)          & 0.0478 (12)         & 0.7134 (8)          & 0.7252 (7)          & 0.3496 (11)         & 0.7273 (6)          & 0.3909 (10)         & 0.7606 (3)          \\
\textbf{arrhythmia2}      & 0.3925 (9)          & 0.463 (4)           & 0.1833 (10)         & 0.4664 (2)          & 0.4239 (7)          & 0.0396 (12)         & 0.3949 (8)          & 0.4323 (5)          & 0.1833 (10)         & 0.4269 (6)          & \textbf{0.6772 (1)} & 0.4664 (2)          \\
\textbf{breastw}          & 0.2822 (12)         & 0.9695 (3)          & 0.9716 (2)          & 0.9684 (4)          & 0.9564 (7)          & 0.3742 (11)         & 0.9504 (8)          & 0.9615 (6)          & 0.817 (9)           & \textbf{0.9741 (1)} & 0.8022 (10)         & 0.9632 (5)          \\
\textbf{Cardio} & 0.2802 (10)         & 0.4454 (4)          & 0.4418 (5)          & 0.4187 (7)          & 0.3299 (9)          & 0.2313 (12)         & \textbf{0.5306 (1)} & 0.2325 (11)         & 0.4989 (2)          & 0.4785 (3)          & 0.3611 (8)          & 0.4233 (6)          \\
\textbf{comm}             & 0.1104 (2)          & 0.0418 (6)          & 0.0486 (5)          & 0.0251 (9)          & 0.0397 (7)          & \textbf{0.6461 (1)} & 0.073 (3)           & 0.0289 (8)          & 0.01 (12)           & 0.0153 (10)         & 0.0153 (10)         & 0.0509 (4)          \\
\textbf{concrete}         & 0.0951 (3)          & 0.0502 (5)          & 0.0153 (12)         & 0.0455 (9)          & 0.0412 (10)         & \textbf{0.1347 (1)} & 0.0493 (6)          & 0.0217 (11)         & 0.0471 (8)          & 0.0493 (6)          & \textbf{0.1347 (1)} & 0.0508 (4)          \\
\textbf{fault}            & 0.2064 (10)         & 0.4269 (4)          & 0.2114 (8)          & 0.4294 (3)          & 0.4364 (2)          & 0.1297 (12)         & 0.2136 (7)          & \textbf{0.4378 (1)} & 0.3838 (6)          & 0.1458 (11)         & 0.2114 (8)          & 0.389 (5)           \\
\textbf{gas}              & 0.0265 (2)          & 0.0017 (4)          & 0.0016 (5)          & 0.0016 (5)          & 0.0018 (3)          & \textbf{0.1382 (1)} & 0.001 (8)           & 0.0009 (10)         & 0.0009 (10)         & 0.001 (8)           & 0 (12)              & 0.0016 (5)          \\
\textbf{glass}            & 0.1388 (2)          & 0.0944 (8)          & 0.068 (11)          & 0.1033 (7)          & 0.1268 (4)          & 0.1318 (3)          & \textbf{0.1393 (1)} & 0.0812 (10)         & 0.0411 (12)         & 0.1183 (6)          & 0.0936 (9)          & 0.1231 (5)          \\
\textbf{Glass2}           & 0.1436 (7)          & 0.2108 (2)          & 0.2066 (3)          & 0.1506 (5)          & 0.1506 (5)          & 0.1295 (9)          & 0.0916 (11)         & 0.0837 (12)         & \textbf{0.2301 (1)} & 0.1115 (10)         & 0.1301 (8)          & 0.2034 (4)          \\
\textbf{HeartDisease}     & 0.4804 (9)          & 0.5306 (5)          & \textbf{0.5646 (1)} & 0.5412 (3)          & 0.5479 (2)          & 0.0517 (12)         & 0.5276 (6)          & 0.5172 (8)          & 0.4604 (11)         & 0.4667 (10)         & 0.5317 (4)          & 0.5204 (7)          \\
\textbf{Hepatitis}        & 0 (11)              & 0.2388 (8)          & 0.3008 (2)          & 0.2527 (5)          & 0.2501 (6)          & 0.2842 (3)          & 0.259 (4)           & 0.2407 (7)          & 0.2012 (10)         & 0 (11)              & 0.2388 (8)          & \textbf{0.329 (1)}  \\
\textbf{imgseg}           & 0.1062 (12)         & 0.3506 (5)          & 0.3635 (3)          & 0.3485 (7)          & 0.3498 (6)          & \textbf{0.4699 (1)} & 0.2688 (8)          & 0.3552 (4)          & 0.1766 (11)         & 0.1844 (9)          & 0.1844 (9)          & 0.3742 (2)          \\
\textbf{InternetAds}      & 0.2557 (10)         & \textbf{0.5101 (1)} & 0.4136 (5)          & 0.4431 (2)          & 0.3385 (8)          & 0.0097 (12)         & 0.419 (4)           & 0.3858 (6)          & 0.3714 (7)          & 0.2173 (11)         & 0.3288 (9)          & 0.4431 (2)          \\
\textbf{ionosphere}       & 0.7949 (4)          & 0.81 (2)            & 0.357 (10)          & 0.7866 (5)          & 0.3474 (11)         & 0.2536 (12)         & 0.81 (2)            & 0.7858 (6)          & 0.565 (9)           & 0.6354 (8)          & 0.6885 (7)          & \textbf{0.8316 (1)} \\
\textbf{landsat}          & 0.1561 (2)          & 0.1291 (9)          & 0.0941 (10)         & 0.1325 (6)          & 0.1314 (8)          & \textbf{0.1628 (1)} & 0.1365 (3)          & 0.1325 (6)          & 0.0941 (10)         & 0.1326 (5)          & 0.0941 (10)         & 0.1336 (4)          \\
\textbf{letter}           & 0.4889 (3)          & 0.0866 (10)         & 0.547 (2)           & 0.0811 (11)         & 0.201 (6)           & \textbf{0.6958 (1)} & 0.1194 (8)          & 0.0951 (9)          & 0.0592 (12)         & 0.3901 (4)          & 0.1682 (7)          & 0.3658 (5)          \\
\textbf{letter}           & \textbf{0.2038 (1)} & 0.0967 (10)         & 0.1606 (2)          & 0.0972 (8)          & 0.0981 (7)          & 0.0073 (11)         & 0.1355 (4)          & 0.0983 (6)          & 0.1554 (3)          & 0.0969 (9)          & 0.0073 (11)         & 0.1193 (5)          \\
\textbf{lympho}           & 0.7817 (8)          & \textbf{1 (1)}      & 0.8968 (7)          & 0.9762 (4)          & 0.6762 (10)         & 0.2054 (12)         & 0.9333 (6)          & 0.9444 (5)          & 0.753 (9)           & \textbf{1 (1)}      & 0.6471 (11)         & \textbf{1 (1)}      \\
\textbf{magic}            & 0.1143 (8)          & 0.1516 (4)          & 0.1219 (7)          & 0.1459 (5)          & 0.1568 (2)          & \textbf{0.2403 (1)} & 0.1104 (9)          & 0.1352 (6)          & 0.0866 (11)         & 0.0479 (12)         & 0.1096 (10)         & 0.155 (3)           \\
\textbf{mammography}      & 0.0793 (11)         & 0.2178 (2)          & 0.1206 (10)         & 0.1783 (4)          & 0.1744 (6)          & 0.0229 (12)         & 0.1609 (8)          & 0.1759 (5)          & \textbf{0.3414 (1)} & 0.1537 (9)          & 0.1692 (7)          & 0.2033 (3)          \\
\textbf{mnist}            & 0.2211 (7)          & 0.2435 (4)          & 0.1589 (9)          & 0.2421 (5)          & 0.1096 (10)         & 0.0982 (11)         & 0.3418 (2)          & 0.2635 (3)          & 0.0785 (12)         & 0.1787 (8)          & 0.2333 (6)          & \textbf{0.4136 (1)} \\
\textbf{musk}             & 0.0662 (12)         & 0.9147 (7)          & 0.9994 (3)          & 0.9964 (6)          & 0.9996 (2)          & 0.1654 (11)         & 0.4462 (10)         & 0.8162 (8)          & 0.9994 (3)          & \textbf{1 (1)}      & 0.6592 (9)          & 0.9992 (5)          \\
\textbf{opt.digits}       & 0.0764 (2)          & 0.0673 (3)          & 0.0609 (8)          & 0.0664 (5)          & 0.0664 (5)          & 0.0607 (9)          & 0.0673 (3)          & 0.0588 (11)         & 0.0554 (12)         & 0.0592 (10)         & 0.0619 (7)          & \textbf{0.0822 (1)} \\
\textbf{optdigits}        & 0.0321 (8)          & 0.0449 (6)          & 0.0222 (10)         & \textbf{0.0639 (1)} & 0.0433 (7)          & 0.0619 (2)          & 0.0274 (9)          & 0.0617 (3)          & 0.0222 (10)         & 0.0515 (5)          & 0.053 (4)           & 0.0219 (12)         \\
\textbf{pageb}            & 0.3763 (8)          & 0.458 (5)           & 0.3615 (9)          & 0.4594 (3)          & 0.4377 (7)          & 0.357 (10)          & 0.4594 (3)          & \textbf{0.4813 (1)} & 0.1635 (11)         & 0.1092 (12)         & \textbf{0.4813 (1)} & 0.458 (5)           \\
\textbf{PageBlocks}       & 0.2861 (9)          & 0.495 (2)           & 0.2654 (10)         & 0.4596 (5)          & 0.4767 (4)          & 0.2333 (11)         & \textbf{0.5039 (1)} & 0.4787 (3)          & 0.2288 (12)         & 0.4202 (7)          & 0.3944 (8)          & 0.4538 (6)          \\
\textbf{particle}         & 0.0633 (8)          & 0.0724 (4)          & 0.0564 (9)          & 0.0841 (2)          & 0.0668 (7)          & \textbf{0.2732 (1)} & 0.0671 (6)          & 0.0519 (10)         & 0.0439 (11)         & 0.0677 (5)          & 0.0439 (11)         & 0.0768 (3)          \\
\textbf{PenDigits}        & 0.0099 (5)          & 0.0048 (7)          & 0.094 (3)           & 0.005 (6)           & 0.0031 (9)          & \textbf{0.5355 (1)} & 0.016 (4)           & 0.0015 (10)         & 0.0012 (11)         & 0.0012 (11)         & 0.1661 (2)          & 0.0045 (8)          \\
\textbf{pendigits2}       & 0.0528 (9)          & 0.2013 (4)          & 0.1143 (5)          & \textbf{0.2843 (1)} & 0.0892 (6)          & 0.0127 (12)         & 0.0789 (7)          & 0.263 (2)           & 0.0414 (10)         & 0.063 (8)           & 0.0134 (11)         & 0.2286 (3)          \\
\textbf{pima}             & 0.4239 (10)         & 0.5139 (2)          & 0.4044 (11)         & 0.4674 (6)          & 0.4674 (6)          & 0.1632 (12)         & 0.4504 (8)          & 0.5073 (3)          & \textbf{0.5586 (1)} & 0.4499 (9)          & 0.4678 (5)          & 0.505 (4)           \\
\textbf{Pima2}            & 0.4606 (8)          & 0.5175 (4)          & 0.3471 (11)         & 0.4874 (6)          & 0.4467 (9)          & 0.0511 (12)         & 0.5314 (2)          & 0.5097 (5)          & 0.5254 (3)          & 0.4446 (10)         & 0.4699 (7)          & \textbf{0.5434 (1)} \\
\textbf{satellite}        & 0.3647 (11)         & \textbf{0.6745 (1)} & 0.6131 (5)          & 0.6315 (4)          & 0.5931 (7)          & 0.5675 (10)         & 0.6082 (6)          & 0.6697 (2)          & 0.3571 (12)         & 0.5757 (8)          & 0.5729 (9)          & 0.6515 (3)          \\
\textbf{satimage-2}       & 0.0332 (10)         & 0.9217 (3)          & 0.7552 (8)          & \textbf{0.9303 (1)} & 0.8964 (5)          & 0.0241 (11)         & 0.8457 (7)          & 0.9167 (4)          & 0.0096 (12)         & 0.8479 (6)          & 0.6211 (9)          & 0.9296 (2)          \\
\textbf{shuttle}          & 0.2052 (11)         & 0.2249 (5)          & 0.3363 (2)          & 0.2282 (4)          & 0.2232 (7)          & \textbf{0.3736 (1)} & 0.2249 (5)          & 0.2413 (3)          & 0.1146 (12)         & 0.2209 (9)          & 0.2232 (7)          & 0.2143 (10)         \\
\textbf{shuttle2}         & 0.0981 (9)          & 0.972 (3)           & 0 (12)              & \textbf{0.9724 (1)} & \textbf{0.9724 (1)} & 0.4726 (5)          & 0.1199 (8)          & 0.968 (4)           & 0.3588 (6)          & 0.0426 (10)         & 0.1532 (7)          & 0.0418 (11)         \\
\textbf{Shuttle3}         & 0.3512 (2)          & 0.0806 (6)          & 0.2628 (4)          & 0.0648 (8)          & 0.0715 (7)          & \textbf{0.5834 (1)} & 0.047 (11)          & 0.0479 (10)         & 0.0126 (12)         & 0.1266 (5)          & 0.0648 (8)          & 0.3481 (3)          \\
\textbf{skin}             & 0.1102 (4)          & 0.0972 (8)          & \textbf{0.1538 (1)} & 0.1003 (7)          & 0.1056 (5)          & 0.066 (12)          & 0.1156 (3)          & 0.0761 (9)          & \textbf{0.1538 (1)} & 0.076 (11)          & 0.0761 (9)          & 0.1007 (6)          \\
\textbf{smtp\_n}          & 0.0012 (10)         & 0.0046 (7)          & 0.0035 (8)          & 0.0074 (5)          & 0.0087 (4)          & \textbf{0.6709 (1)} & 0.0048 (6)          & 0.0455 (3)          & 0 (11)              & 0 (11)              & 0.0035 (8)          & 0.2227 (2)          \\
\textbf{spambase}         & 0.012 (9)           & 0.0238 (4)          & 0.0036 (10)         & 0.0228 (5)          & 0.0248 (3)          & 0.3843 (2)          & 0.0176 (8)          & 0.0189 (6)          & 0.0036 (10)         & 0.0036 (10)         & \textbf{0.4243 (1)} & 0.0189 (6)          \\
\textbf{SpamBase2}        & 0.3516 (10)         & 0.4666 (7)          & 0.3029 (11)         & 0.4654 (8)          & 0.4842 (6)          & 0.5261 (4)          & \textbf{0.5369 (1)} & 0.5283 (2)          & 0.3015 (12)         & 0.3802 (9)          & 0.5283 (2)          & 0.505 (5)           \\
\textbf{speech}           & 0.0284 (3)          & 0.0193 (9)          & 0.0324 (2)          & 0.0246 (5)          & 0.0147 (11)         & 0.024 (6)           & 0.0213 (8)          & 0.0158 (10)         & 0.0134 (12)         & \textbf{0.0987 (1)} & 0.023 (7)           & 0.0281 (4)          \\
\textbf{Stamps}           & 0.1453 (12)         & 0.3326 (2)          & 0.1569 (10)         & 0.2957 (5)          & 0.2856 (7)          & 0.1532 (11)         & \textbf{0.3383 (1)} & 0.2948 (6)          & 0.2411 (9)          & 0.3006 (3)          & 0.2812 (8)          & 0.2982 (4)          \\
\textbf{synthetic}        & \textbf{0.1552 (1)} & 0.1192 (5)          & 0.1477 (2)          & 0.1197 (4)          & 0.113 (7)           & 0.0938 (11)         & 0.0986 (9)          & 0.0974 (10)         & 0.1032 (8)          & 0 (12)              & 0.1192 (5)          & 0.1464 (3)          \\
\textbf{thyroid}          & 0.0356 (11)         & 0.6295 (6)          & \textbf{0.7751 (1)} & 0.6054 (7)          & 0.3558 (8)          & 0.0603 (9)          & 0.732 (3)           & 0.697 (4)           & 0.0265 (12)         & \textbf{0.7751 (1)} & 0.0603 (9)          & 0.6677 (5)          \\
\textbf{vertebral}        & 0.1135 (3)          & 0.0917 (9)          & 0.0918 (8)          & 0.0925 (7)          & 0.0996 (6)          & \textbf{0.6885 (1)} & 0.0877 (11)         & 0.0856 (12)         & 0.0998 (5)          & 0.1175 (2)          & 0.1052 (4)          & 0.0887 (10)         \\
\textbf{vowels}           & 0.3005 (4)          & 0.1679 (5)          & 0.3438 (3)          & 0.0997 (8)          & 0.0754 (9)          & 0.1031 (7)          & \textbf{0.8041 (1)} & 0.1452 (6)          & 0.0193 (11)         & 0.0279 (10)         & 0.0193 (11)         & 0.6355 (2)          \\
\textbf{wave}             & 0.0479 (3)          & 0.0128 (7)          & 0.01 (10)           & 0.0115 (9)          & 0.0124 (8)          & \textbf{0.3835 (1)} & 0.0479 (3)          & 0.0147 (6)          & 0.0557 (2)          & 0.0076 (12)         & 0.01 (10)           & 0.0339 (5)          \\
\textbf{Waveform}         & 0.0834 (4)          & 0.0593 (8)          & 0.1306 (3)          & 0.0543 (9)          & 0.0537 (10)         & \textbf{0.4882 (1)} & 0.0478 (12)         & 0.051 (11)          & 0.2756 (2)          & 0.0596 (7)          & 0.0659 (5)          & 0.064 (6)           \\
\textbf{wbc}              & 0.5394 (8)          & 0.5783 (6)          & \textbf{0.6497 (1)} & 0.594 (2)           & 0.5226 (9)          & 0.0503 (12)         & 0.3401 (10)         & 0.5736 (7)          & 0.0989 (11)         & 0.5853 (5)          & 0.594 (2)           & 0.594 (2)           \\
\textbf{WBC2}             & 0.1038 (9)          & \textbf{0.8643 (1)} & 0.0283 (10)         & 0.8562 (2)          & 0.7213 (3)          & 0.2639 (8)          & 0.0283 (10)         & 0.3856 (7)          & 0.0283 (10)         & 0.6586 (5)          & 0.6575 (6)          & 0.6613 (4)          \\
\textbf{WDBC}             & \textbf{0.729 (1)}  & 0.6773 (4)          & 0.6699 (5)          & 0.6449 (8)          & 0.6962 (3)          & 0.4283 (11)         & 0.7252 (2)          & 0.6579 (7)          & 0.1268 (12)         & 0.6202 (9)          & 0.5957 (10)         & 0.6622 (6)          \\
\textbf{Wilt}             & 0.1001 (2)          & 0.0485 (7)          & 0.0602 (5)          & 0.0402 (11)         & 0.0446 (9)          & \textbf{0.6381 (1)} & 0.0653 (3)          & 0.0485 (7)          & 0.0446 (9)          & 0.0644 (4)          & 0.0495 (6)          & 0.0402 (11)         \\
\textbf{wine}             & 0.0066 (12)         & 0.0091 (8)          & 0.0127 (3)          & 0.0107 (4)          & 0.0092 (7)          & \textbf{0.644 (1)}  & 0.0085 (9)          & 0.0101 (6)          & 0.0075 (10)         & 0.0075 (10)         & 0.2477 (2)          & 0.0103 (5)          \\
\textbf{wine2}            & 0.176 (8)           & 0.2442 (4)          & 0.1108 (9)          & 0.2008 (5)          & 0.3074 (3)          & 0.3818 (2)          & 0.0745 (10)         & 0.1832 (6)          & 0.0536 (12)         & 0.0571 (11)         & 0.1832 (6)          & \textbf{0.8697 (1)} \\
\textbf{WPBC}             & 0.2317 (9)          & 0.2311 (10)         & 0.2363 (5)          & 0.223 (12)          & 0.2487 (2)          & \textbf{0.3333 (1)} & 0.2487 (2)          & 0.2318 (8)          & 0.2277 (11)         & 0.2358 (6)          & 0.2329 (7)          & 0.2465 (4)          \\
\textbf{yearp}            & \textbf{0.4978 (1)} & 0.4922 (4)          & 0.4937 (2)          & 0.4921 (5)          & 0.4805 (10)         & 0.2076 (12)         & 0.4877 (9)          & 0.4901 (7)          & 0.4758 (11)         & 0.4905 (6)          & 0.4937 (2)          & 0.4881 (8)          \\
\textbf{yeast}            & 0.0596 (7)          & 0.058 (9)           & 0.0561 (12)         & 0.0586 (8)          & 0.064 (5)           & \textbf{0.1227 (1)} & 0.0675 (2)          & 0.0637 (6)          & 0.0576 (10)         & 0.0575 (11)         & 0.0675 (2)          & 0.0672 (4)          \\
\midrule
\textbf{Average}          & 0.2154 (6.87)       & 0.3197 (5.08)       & 0.2689 (6.44)       & 0.3137 (5.42)       & 0.2892 (6.13)       & 0.2704 (6.53)       & 0.2814 (5.73)       & 0.2981 (6.6)        & 0.1946 (8.97)       & 0.2594 (7.68)       & 0.2582 (6.9)        & \textbf{0.3382 (4.53)}       \\
\textbf{STD}              & 0.2027              & 0.2967              & 0.2602              & 0.2966              & 0.2712              & 0.221               & 0.2691              & 0.2839              & 0.2199              & 0.2813              & 0.2268              & 0.2882        \\     

\bottomrule
\end{tabular}}
    \normalsize
\end{table*}
\begin{table*}[!htp]
\scriptsize
\caption{
Pairwise statistical test results between \method and baselines by Wilcoxon signed rank test in ST. Statistically better method shown in \textbf{bold} (both marked \textbf{bold} if no significance). \method related pairs are surrounded by rectangles. \method achieves the highest MAP and best average rank, and statistically significantly better than all baselines except iForest. 
}\label{table:st_pairs_full}
\scalebox{0.92}{
\begin{tabular}{lll}
\centering
\begin{tabular}{ll|l}
\toprule
\textbf{Method 1}         & \textbf{Method 2}           & \textbf{p-value} \\
\midrule
LOF (0.2154)                & \textbf{IForest (0.3197)}   & 0.0032           \\
\textbf{LOF (0.2154)}       & \textbf{ME (0.2689)}        & 0.5258           \\
LOF (0.2154)                & \textbf{GB (0.3137)}        & 0.01             \\
\textbf{LOF (0.2154)}       & \textbf{ISAC (0.2892)}      & 0.1016           \\
\textbf{LOF (0.2154)}       & \textbf{AS (0.2704)}        & 0.1683           \\
\textbf{LOF (0.2154)}       & \textbf{SS (0.2814)}        & 0.0575           \\
LOF (0.2154)                & \textbf{ALORS (0.2981)}     & 0.0461           \\
LOF (0.2154)                & \textbf{MetaOD\_C (0.1946)} & 0.0345           \\
\textbf{LOF (0.2154)}       & \textbf{MetaOD\_F (0.2594)} & 0.6176           \\
\textbf{LOF (0.2154)}       & \textbf{RS (0.2582)}        & 0.1735           \\
\marktopleft{c1}LOF (0.2154)                & \textbf{MetaOD (0.3382)} \markbottomright{c1}   & 0.0001           \\
\textbf{IForest (0.3197)}   & ME (0.2689)                 & 0.0484           \\
\textbf{IForest (0.3197)}   & GB (0.3137)                 & 0.047            \\
\textbf{IForest (0.3197)}   & ISAC (0.2892)               & 0.0127           \\
\textbf{IForest (0.3197)}   & \textbf{AS (0.2704)}        & 0.3714           \\
\textbf{IForest (0.3197)}   & \textbf{SS (0.2814)}        & 0.1942           \\
\textbf{IForest (0.3197)}   & ALORS (0.2981)              & 0.0012           \\
\textbf{IForest (0.3197)}   & MetaOD\_C (0.1946)          & 0.0001                \\
\textbf{IForest (0.3197)}   & MetaOD\_F (0.2594)          & 0.0001           \\
\textbf{IForest (0.3197)}   & RS (0.2582)                 & 0.0012           \\
\marktopleft{c2}\textbf{IForest (0.3197)}   & \textbf{MetaOD (0.3382)}\markbottomright{c2}     & 0.1129           \\
\textbf{ME (0.2689)}        & \textbf{GB (0.3137)}        & 0.1002           \\
\bottomrule
\end{tabular}
     & 
\begin{tabular}{ll|l}
\toprule
\textbf{Method 1}         & \textbf{Method 2}           & \textbf{p-value} \\
\midrule

\textbf{ME (0.2689)}        & \textbf{ISAC (0.2892)}      & 0.6817           \\
\textbf{ME (0.2689)}        & \textbf{AS (0.2704)}        & 0.9525           \\
\textbf{ME (0.2689)}        & \textbf{SS (0.2814)}        & 0.2345           \\
\textbf{ME (0.2689)}        & \textbf{ALORS (0.2981)}     & 0.5031           \\
\textbf{ME (0.2689)}        & MetaOD\_C (0.1946)          & 0.0026           \\
\textbf{ME (0.2689)}        & \textbf{MetaOD\_F (0.2594)} & 0.2389           \\
\textbf{ME (0.2689)}        & \textbf{RS (0.2582)}        & 0.3672           \\
\marktopleft{c3}ME (0.2689)                 & \textbf{MetaOD (0.3382)}\markbottomright{c3}    & 0.0001           \\
\textbf{GB (0.3137)}        & \textbf{ISAC (0.2892)}      & 0.0849           \\
\textbf{GB (0.3137)}        & \textbf{AS (0.2704)}        & 0.4468           \\
\textbf{GB (0.3137)}        & \textbf{SS (0.2814)}        & 0.5412           \\
\textbf{GB (0.3137)}        & \textbf{ALORS (0.2981)}     & 0.1002           \\
\textbf{GB (0.3137)}        & MetaOD\_C (0.1946)          & 0.0001                \\
\textbf{GB (0.3137)}        & MetaOD\_F (0.2594)          & 0.0001           \\
\textbf{GB (0.3137)}        & RS (0.2582)                 & 0.0017           \\
\marktopleft{c4}GB (0.3137)                 & \textbf{MetaOD (0.3382)}  \markbottomright{c4}  & 0.0030            \\
\textbf{ISAC (0.2892)}      & \textbf{AS (0.2704)}        & 0.7233           \\
\textbf{ISAC (0.2892)}      & \textbf{SS (0.2814)}        & 0.9513           \\
\textbf{ISAC (0.2892)}      & \textbf{ALORS (0.2981)}     & 0.9357           \\
\textbf{ISAC (0.2892)}      & MetaOD\_C (0.1946)          & 0.0003           \\
\textbf{ISAC (0.2892)}      & MetaOD\_F (0.2594)          & 0.0159           \\
\textbf{ISAC (0.2892)}      & RS (0.2582)                 & 0.0344           \\
\bottomrule
\end{tabular}
&
\begin{tabular}{ll|l}
\toprule
\textbf{Method 1}         & \textbf{Method 2}           & \textbf{p-value} \\
\midrule

\marktopleft{c5}ISAC (0.2892)               & \textbf{MetaOD (0.3382)}\markbottomright{c5}    & 0.0006           \\
\textbf{AS (0.2704)}        & \textbf{SS (0.2814)}        & 0.8471           \\
\textbf{AS (0.2704)}        & \textbf{ALORS (0.2981)}     & 0.5725           \\
\textbf{AS (0.2704)}        & \textbf{MetaOD\_C (0.1946)} & 0.0667           \\
\textbf{AS (0.2704)}        & \textbf{MetaOD\_F (0.2594)} & 0.6112           \\
\textbf{AS (0.2704)}        & \textbf{RS (0.2582)}        & 0.8622           \\
\marktopleft{c6}\textbf{AS (0.2704)}        & \textbf{MetaOD (0.3382)}\markbottomright{c6}    & 0.0009           \\
\textbf{SS (0.2814)}        & \textbf{ALORS (0.2981)}     & 0.7604           \\
\textbf{SS (0.2814)}        & MetaOD\_C (0.1946)          & 0.0001           \\
\textbf{SS (0.2814)}        & MetaOD\_F (0.2594)          & 0.0069           \\
\textbf{SS (0.2814)}        & \textbf{RS (0.2582)}        & 0.1419           \\
\marktopleft{c7}SS (0.2814)                 & \textbf{MetaOD (0.3382)}\markbottomright{c7}   & 0.0190            \\
\textbf{ALORS (0.2981)}     & MetaOD\_C (0.1946)          & 0.0001           \\
\textbf{ALORS (0.2981)}     & MetaOD\_F (0.2594)          & 0.0280            \\
\textbf{ALORS (0.2981)}     & \textbf{RS (0.2582)}        & 0.0524           \\
\marktopleft{c8}ALORS (0.2981)              & \textbf{MetaOD (0.3382)}\markbottomright{c8}    & 0.0001           \\
MetaOD\_C (0.1946)          & \textbf{MetaOD\_F (0.2594)} & 0.0342           \\
MetaOD\_C (0.1946)          & \textbf{RS (0.2582)}        & 0.0115           \\
\marktopleft{c9}MetaOD\_C (0.1946)          & \textbf{MetaOD (0.3382)}\markbottomright{c9}    & 0.0001                \\
\textbf{MetaOD\_F (0.2594)} & \textbf{RS (0.2582)}        & 0.4616           \\
\marktopleft{c10}MetaOD\_F (0.2594)          & \textbf{MetaOD (0.3382)}\markbottomright{c10}    & 0.0001                \\
\marktopleft{c11}RS (0.2582)                 & \textbf{MetaOD (0.3382)}\markbottomright{c11}    & 0.0001           \\ 
\bottomrule
\end{tabular}
\end{tabular}}
\vspace{0.05in}
\end{table*}

\subsection{Runtime Analysis}
\label{appendix:runtime}

\begin{figure}[!t]
\centering
    \includegraphics[scale=0.41]{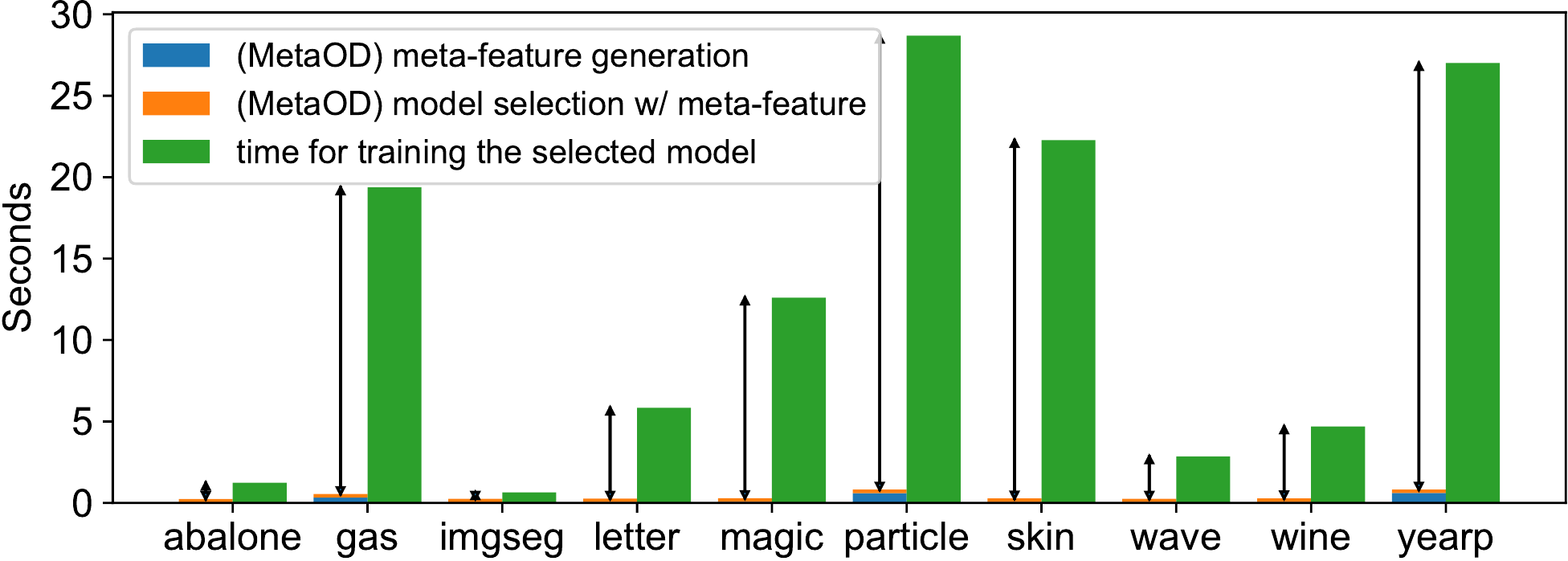}
    \vspace{-0.1in}
    \caption{\label{fig:metaod_time}
    Time for \method  vs. training of the selected model (on 10 largest datasets in POC). \method  incurs only negligible overhead (diff. shown w/ black arrows).
    } 
    \vspace{0.05in}
\end{figure}

As discussed in \S \ref{sec:exp-runtime}, the runtime overhead of \method (i.e., meta-feature generation and performance estimation) is negligible in comparison to the training time of the selected model. Fig. \ref{fig:metaod_time} corroborates the statement by showing the comparison on the 10 largest datasets in POC.

Notably, meta-feature extraction may be trivially parallelized whereas the model selection is even faster, e.g., using SUOD \cite{zhao2021suod}, effectively taking constant time (See \S \ref{subsubsec:prediction_model_selection}).

\label{sec:appendix}




\end{document}